\documentclass{article}

 \PassOptionsToPackage{numbers, sort&compress}{natbib}

\usepackage[preprint, main]{neurips_2026}

\usepackage[utf8]{inputenc}
\usepackage[T1]{fontenc}
\usepackage{hyperref}
\hypersetup{colorlinks=true,linkcolor=blue,citecolor=blue,urlcolor=blue}
\usepackage{url}
\usepackage{booktabs}
\usepackage{amsfonts}
\usepackage{amsmath}
\usepackage{mathtools}
\usepackage{microtype}
\usepackage{xcolor}
\usepackage{graphicx}
\usepackage{multirow}
\usepackage{enumitem}
\usepackage{array}
\usepackage{makecell}
\usepackage{tikz}
\usepackage{fontawesome}
\usetikzlibrary{arrows.meta,positioning,fit,calc,shapes.misc}
\usepackage{wrapfig}
\usepackage{listings}
\lstdefinestyle{promptstyle}{
  basicstyle=\ttfamily\scriptsize,
  columns=fullflexible,
  keepspaces=true,
  breaklines=true,
  breakatwhitespace=false,
  frame=single,
  framerule=0.25pt,
  rulecolor=\color{black!35},
  aboveskip=4pt,
  belowskip=6pt
}

\newcommand{\pre}{h_{\mathrm{pre}}}
\newcommand{\post}{h_{\mathrm{post}}}
\newcommand{\conn}{\mathcal{C}}

\newcommand{\model}[1]{#1}
\newcommand{\bench}{\model{TRACE-Edit}}
\newcommand{\uhid}{\model{U-Hidden}}
\newcommand{\uquery}{\model{U-Query}}
\newcommand{\kiwi}{\model{Kiwi-Edit}}
\newcommand{\wanctrl}{\model{Wan-Query}}

\newcommand{\tightcaption}[1]{\caption{#1}}

\title{What Semantics Survive the Connector?\\
Diagnosing VLM-to-DiT Alignment in Video Editing}

\author{%
  Hangyu Lin\textsuperscript{*}\qquad
  Chao Wen\textsuperscript{*}\qquad
  Chengming Xu\textsuperscript{\faEnvelopeO}\qquad
  JianXiong Gao\\
  \textbf{Jiangning Zhang}\qquad
  \textbf{Xiaobin Hu}\qquad
  \textbf{Yanwei Fu}\\
  HKUST\qquad FDU\qquad ZJU\qquad NUS
}
\begin{document}
\maketitle

\begin{abstract}
Flow matching based video generative models have been increasingly relying on prepended Vision-Language Models (VLMs) to handle complex, instruction-based video editing. The prevailing assumption underlying this paradigm is that a connector module can seamlessly align the VLM's rich multi-modal reasoning with the original text embedding space of DiTs. However, we hypothesize that this alignment acts as a severe semantic bottleneck, degrading fine-grained structural variables. Verifying this is challenging, as end-to-end evaluations conflate alignment failures with generation errors, and natural datasets lack disentangled annotations. To rigorously investigate this, we propose a controlled data processing pipeline based on video composition that results in \bench{}, a diagnostic dataset focusing on relation-based editing. Leveraging this dataset, we propose a comprehensive diagnostic protocol to analyze two important designs of meta-query and connector in the existing video editing models. Systematic evaluation of four representative model cases reveals that fine-grained structural semantics can be severely degraded during alignment. Our findings overturn the assumption of lossless semantic transfer, identifying the VLM-to-DiT alignment as a major bottleneck and providing a new diagnostic foundation for future multi-modal alignment architectures.
\end{abstract}

\section{Introduction \label{sec:intro}}

Modern video generative models, such as Wan~\citep{wan2025}, HunyuanVideo~\citep{hunyuanvideo2024}, and LTX~\citep{hacohen2026ltx}, have demonstrated remarkable capabilities in synthesizing high-quality motion and appearance. However, these models cannot natively be applied to instruction-based video editing, demanding complex multi-modal reasoning, such as identifying \emph{where} to edit, \emph{what} attribute to apply, and \emph{which} object must serve as a reference. Besides the limitation of pretraining tasks, the text encoders of these models often struggle with such fine-grained multi-modal binding. To this end, recent models such as UniVideo~\citep{univideo2026} and Kiwi-Edit~\citep{kiwiedit2026} have adopted a seemingly straightforward paradigm to replace the original text encoder with a more powerful multi-modal one, such as Qwen2.5VL~\citep{qwen25vl2025}, by leveraging an MLP connector to bridge the latent spaces of these two encoders. Furthermore, these models adopt extra query tokens to enhance the condition from multi-modal instruction. The prevailing assumption underlying such a design is that the connector module can align the VLM's rich semantic understanding with the DiT's original condition space, based on which the query tokens provide further refinement, thereby transferring superior instruction-following capabilities.

\begin{figure}[htbp]
    \centering
    \includegraphics[width=0.9\linewidth]{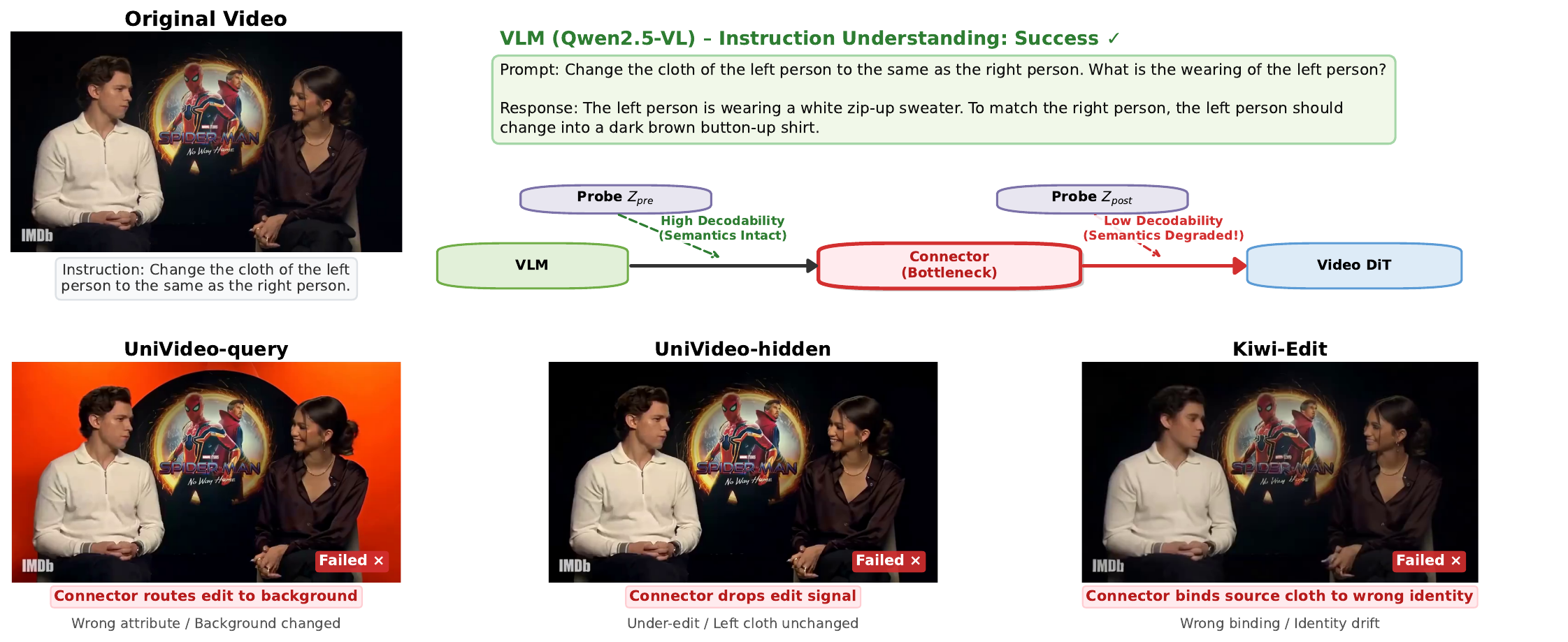} 
    \caption{
        \textbf{The Connector Bottleneck in VLM-driven Video Editing.} Although the VLM accurately comprehends instructions (top), the alignment connector degrades rich structural semantics. This compromised conditioning causes downstream Video DiT failures, including misrouted attributes, dropped signals, and identity drift (bottom).}
    \vspace*{-0.25in}
    \label{fig:teaser}
\end{figure}

Unfortunately, as illustrated in Fig.~\ref{fig:teaser}, while VLMs can successfully understand the edit instruction, the video editing models with the same VLMs fail in different ways such as routing edits to the background, dropping the edit signal entirely, or binding attributes to the wrong identity. It seems aligning the latent space of the strongly performing VLMs to their weaker counterparts may make the post-connector representations fail to preserve critical edit-related semantics and deliver compromised conditions to the DiT. Therefore it leads to an important question: \textit{Will current design result in semantic bottleneck?} Verifying and isolating such bottleneck is challenging, as existing end-to-end video editing benchmarks often conflate these alignment failures with the DiT's own denoising or attention errors, while overlooking the need for disentangled probing.

To rigorously investigate this, we introduce \bench{}, a controlled diagnostic benchmark and evaluation protocol specifically focusing on analyzing the VLM-to-DiT alignment. 
Rather than relying on unstructured natural videos, \bench{} employs a factorized relation-editing protocol. We first synthesize single-object atomic videos with specific attributes, which are verified and filtered with VLMs, then randomly compose them into video grids.
By pairing these composed videos with instructions that specify edited and reference slots, a large-scale synthetic video editing benchmark with structural instructions can be easily built at low costs. 
Crucially, every example is equipped with exact ground-truth labels for slots, attributes, and object roles. 
TRACE-Edit is not meant to replace natural video editing benchmarks, but to isolate hidden failure modes that natural benchmarks conflate. 
Leveraging this benchmark, we propose a diagnostic framework to probe the internal properties of VLM-to-DiT alignment, specifically condition space geometry, semantic accessibility, token-level evidence routing, and attention allocation. Our findings overturn the prevailing assumption that connectors are semantic-preserving adapters. Although coarse intent survives, fine-grained structural semantics severely degrade during alignment (e.g., target-slot decodability drops by \(18.4\%\) in \uquery{}). Crucially, token-route analysis reveals a severe misalignment regarding query tokens: while text tokens retain strong semantics, alignment drastically shifts actual semantic evidence away from query tokens, yet the DiT still allocates dominant attention to them. This disconnect between information routing and attention allocation destroys slot--value--role bindings. Consequently, edit result judgment performed by VLMs confirm that structural misassignments (accounting for up to \(61.9\%\) of errors), rather than conservative under-editing, constitute the primary failure mode, exposing the connector as a major semantic bottleneck.
In summary, our main contributions are:
\begin{enumerate}[leftmargin=*,nosep]
    \item We introduce \bench{}, a controlled relation-editing benchmark that isolates slot, value, and object-role variables via synthetic video composition, providing precise labels for diagnosing structural semantics.
    \item We propose a comprehensive diagnostic protocol utilizing specifically for the VLM-to-DiT alignment.
    \item We present a rigorous empirical study demonstrating that current VLM-to-DiT designs suffer from semantic degradation, thus identifying the connector as a primary bottleneck in video editing models.
\end{enumerate}

\section{Related Work}
\noindent \textbf{VLM-related Video Diffusion Models}
A growing family of unified multi-modal generation systems couples an autoregressive VLM/MLLM with a diffusion or DiT decoder. MetaQueries introduce learnable query tokens as the bridge from MLLM latents to a diffusion generator \citep{metaqueries2025}. Query-Kontext uses multi-modal ``kontext'' tokens to bridge a VLM and diffusion model for image generation and editing \citep{querykontext2025}. BLIP3-o and ILLUME+ explore hybrid multi-modal understanding-generation architectures with diffusion components \citep{blip3o2025,illumeplus2025}, while OmniGen shows that unified image generation can be driven by instruction-like multi-modal prompts \citep{omnigen2024}. Recent video-oriented systems such as UniVideo, Kiwi-Edit, OmniWeaving, InstructX, and VINO extend the trend to video generation and editing by combining VLM-side reasoning with DiT-side synthesis \citep{univideo2026,kiwiedit2026,vino2026,pan2026omniweavingunifiedvideogeneration,mou2025instructxunifiedvisualediting}.

\noindent \textbf{Video Editing Benchmarks}
Recent benchmarks have significantly advanced the evaluation of text-guided generation and editing. Works such as EditVal \citep{basu2023editval}, VBench \citep{huang2024vbench}, and EvalCrafter \citep{liu2024evalcrafter} provide comprehensive evaluation suites for image editing and video generative models. Building on this, newer datasets like EditVerseBench \citep{ju2025editverse} and IVEBench \citep{chen2025ivebench} have further modernized the landscape by introducing complex instruction-guided taxonomies and MLLM-based metrics for video editing. These benchmarks rely on unstructured, in-the-wild videos that lack disentangled annotations for spatial slots and relational bindings. In contrast, \bench{} is explicitly designed as a diagnostic benchmark specifically for probing intermediate features and pinpointing the exact semantic bottlenecks in VLM-to-DiT alignment pathways, a capability absent in prior benchmarks.

\section{The \bench{} Diagnostic Benchmark}
\label{sec:benchmark}

\subsection{Task Formulation and Variables}

\bench{} is specifically designed to make structural editing semantics observable and quantifiable. A single example in our benchmark consists of a short composite video \(x\), a natural language instruction \(I\), and a comprehensive label set:
\[
    y = (y_{\mathrm{attr}}, y_{\mathrm{slot}}, y_{\mathrm{tgt}}, y_{\mathrm{editobj}}, y_{\mathrm{refobj}})
\]
To break this down:
\begin{itemize}
    \item \(y_{\mathrm{attr}}\): The attribute type being edited (color, material, or action).
    \item \(y_{\mathrm{slot}}\): The spatial slot of the object to be edited.
    \item \(y_{\mathrm{tgt}}\): The target value inherited from the reference object.
    \item \(y_{\mathrm{editobj}} / y_{\mathrm{refobj}}\): The names of the edited and reference objects, respectively.
\end{itemize}

\begin{wrapfigure}{r}{0.55\textwidth}
    \vspace{-1.5em}
    \centering
    \includegraphics[width=\linewidth]{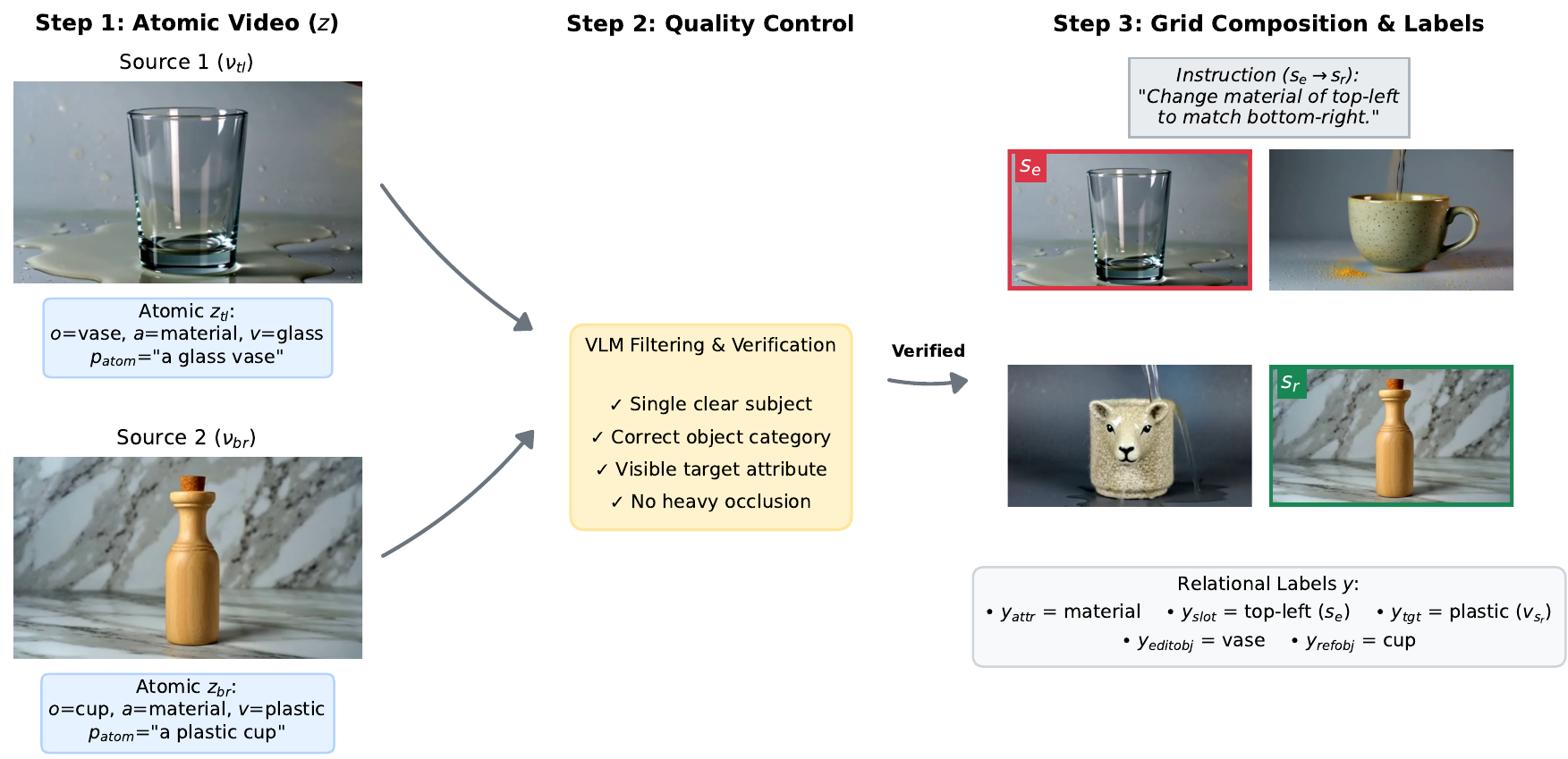} 
    \caption{\textbf{Pipeline for constructing \bench{}.} 
    }
    \label{fig:task_example}
    \vspace{-1em}
\end{wrapfigure}

\noindent \textbf{Relational Labels and Grid Layout.} We adopt a four-slot grid layout (top-left, top-right, bottom-left, bottom-right). This deliberately raises the entropy of the ``where'' decision from a simple binary (left/right) to a four-way choice, while keeping the visual structure clean enough for controlled probing. As illustrated in Fig.~\ref{fig:task_example}, a typical relation instruction asks the editor to modify one cell based on another. Given the instruction \textit{``Change the material of the object in the top-left position to match the object in the bottom-right position,''} the benchmark extracts the structured relational labels \(y\). Specifically, it identifies the attribute to be modified (\(y_{\mathrm{attr}} = \mathrm{material}\)), the spatial location of the target video (\(y_{\mathrm{slot}} = \mathrm{top\text{-}left}\)), and the target attribute value derived from the reference video (\(y_{\mathrm{tgt}} = \mathrm{plastic}\)). Furthermore, it grounds the object categories for both slots, identifying the edited object as \(y_{\mathrm{editobj}} = \mathrm{vase}\) and the reference object as \(y_{\mathrm{refobj}} = \mathrm{cup}\).

\noindent \textbf{Target Formulation.} Formally, let \(S=\{\mathrm{tl},\mathrm{tr},\mathrm{bl},\mathrm{br}\}\) be the slot set. A composite scene contains cell metadata \(G = \{c_s=(o_s, a, v_s, \nu_s): s\in S\}\), where \(o_s\) is the object, \(a\) is the shared attribute type, \(v_s\) is the slot-specific value, and \(\nu_s\) is the underlying atomic video. When we sample an edited slot \(s_e\) and a distinct reference slot \(s_r\), the target label deterministically becomes \(y_{\mathrm{tgt}} = v_{s_r}\) and \(y_{\mathrm{slot}}=s_e\). This strict formulation allows us to evaluate whether a model preserves the exact semantic variables needed for correct editing, rather than merely generating a visually plausible video.

\subsection{Benchmark Construction Pipeline}

To ensure high-quality compositions and precise control, \bench{} is built from the ground up through a rigorous pipeline, starting from single-object videos and culminating in complex relational grids.

\noindent \textbf{Atomic Video Generation.} We generate 81-frame clips at \(480\times832\) resolution for each atomic description using the Wan2.2-A14B text-to-video model \cite{wan2025}. An atomic instance is defined as:
\begin{equation}
    z = (u, o, a, v, p_{\mathrm{atom}}, \nu)
\end{equation}
where \(u\) is a unique identifier, \(o\) is drawn from a 15-object pool, \(a \in \{\mathrm{color}, \mathrm{material}, \mathrm{action}\}\), \(v\) is the attribute value, \(p_{\mathrm{atom}}\) is the generation prompt, and \(\nu\) is the generated video. Our current instantiation features 9 colors, 10 materials, and 9 actions. To facilitate fine-grained generation, the system parameterizes the source videos into these atomic variables \(z\). Returning to the example in Fig.~\ref{fig:task_example}, for the top-left source video, the benchmark identifies the base object \(o = \mathrm{vase}\) and its original material \(v = \mathrm{glass}\), constructing the atomic prompt \(p_{\mathrm{atom}} = \text{``a glass vase''}\). Similarly, for the bottom-right reference video, it extracts \(o = \mathrm{cup}\) and \(v = \mathrm{plastic}\), yielding the prompt \(p_{\mathrm{atom}} = \text{``a plastic cup''}\). These atomic prompts are generated via compositional natural-language templates (e.g., adjectives for color/material, motion phrases for actions) wrapped in strict constraints to ensure a single clear subject and stable camera motion.

\noindent \textbf{Quality Control via VLM Filtering.} Crucially, before any composition occurs, we apply a strict Vision-Language Model (VLM) verification stage. The VLM acts as an automated quality inspector, answering a structured questionnaire to verify: (1) the presence of a single central subject, (2) the correct object category \(o\), and (3) the visibility of the attribute value \(v\). Only videos that pass all checks are admitted into the composition pool. This atomic-level filtering is vital: it guarantees that any failure in the final relational editing task is due to the model's inability to understand relational binding, rather than a hallucination or rendering failure in the underlying source videos.

\noindent \textbf{Grid Composition and Relational Direction.} Verified atomic videos are grouped by attribute type. For a single composite scene, we select one attribute type and draw four distinct values, randomly assigning four corresponding atomic videos to the ordered grid slots \(S\). The videos are spatially resized into tiles, pasted into their respective slots, and temporally synchronized. A key design feature of \bench{} is its sensitivity to \textbf{relational direction}. From a single composite scene, we sample an edited slot \(s_e\) and a reference slot \(s_r\), and generate two directed instructions: one forward (\(s_e \rightarrow s_r\)) and one inverted (\(s_r \rightarrow s_e\)). Both instructions share the exact same visual input but require completely different target values and object roles. A model that merely recognizes the scene without understanding directional instructions will fail this test. Moreover, our benchmark can be easily extended to more complicated version by compositing more video grids, which can be useful for more powerful video editing models in the future.

\begin{wraptable}{r}{0.42\textwidth} 
\centering
\small
\begin{tabular}{lc}
\toprule
\textbf{Quantity} & \textbf{Value} \\
\midrule
Composite videos & \(5{,}524\) \\
Relation examples & \(11{,}048\) \\
Relation directions & \(5{,}524/5{,}524\) \\
Slot labels & \(4\) classes \\
Attribute examples & \(6{,}490/3{,}096/1{,}462\) \\
Semantic labels & \(5\) groups \\
\bottomrule
\end{tabular}
\caption{Summary of \bench{}.} 
\label{tab:bench_stats}
\vspace{-2em}
\end{wraptable}

\noindent \textbf{Instruction Generation and Dataset Scale.} Instructions are instantiated using attribute-specific templates (e.g., matching colors/materials, or following motion trajectories) and include an evaluation query designed to verify the edit, whose ground-truth answer is \(v_{s_r}\). As summarized in Tab.~\ref{tab:bench_stats}, \bench{} operates at a significant scale, containing \(5{,}524\) unique composite videos yielding \(11{,}048\) relation-editing examples. Because each composite video provides both a forward and an inverted instruction, the dataset is perfectly balanced to diagnose true relational comprehension across color (\(6{,}490\) examples), material (\(3{,}096\)), and action (\(1{,}462\)) domains. Detailed object/value pools, atomic prompt templates, VLM verifier prompts, and relation-row metadata are provided in Append.~\ref{app:trace_edit_details}.
\section{Experimental Protocol}
\label{sec:protocol}

Our experiments are organized as connector diagnostics rather than a video-editing leaderboard. For each evaluated model, we compare the VLM representation before and after connector alignment, and ask whether the task-critical information defined by \bench{} remains accessible. This framing isolates the connector as a semantic transfer module: the goal is not to rank editors by final video quality, but to test whether the representation consumed by the video DiT still exposes enough condition-related information needed for the edit.

Tab.~\ref{tab:interfaces} summarizes the four models utilized in our experiments. \uhid{} and \uquery{} represent UniVideo based on HunyuanVideo and transfer editing condition through vision-and-text and text-and-query routes, respectively. \kiwi{} represents a query-based video editor built on Wan2.2 5B. \wanctrl{} is a model trained by us, which shares the same condition embedding mechanism as \uquery{} to embed the editing condition, but instead adopts Wan2.1 14B as DiT backbone and uses the condition embedding through cross-attention. We follow UniVideo and Kiwi-Edit to perform a multi-stage training strategy, i.e. (1) train the connector with T2V and I2V data, (2) train connector and DiT parameters with T2V and I2V data, and (3) train connector and DiT parameters with video editing data.
Although differences in training data preclude a fair comparison of final editing quality, including \wanctrl{} allows us to isolate the impact of connector and query designs, ruling out the confounding factor of a weak DiT backbone.

\begin{wraptable}{r}{0.45\textwidth}
\vspace{-1.5em} % 微调表格与上方文字的间距
\centering
\small
\setlength{\tabcolsep}{4pt}
\begin{tabular}{lcc}
\toprule
\textbf{Model}  & \textbf{DiT backbone} & \textbf{Condition route} \\
\midrule
\uhid{}  & \multirow{2}{*}{HunyuanVideo} & text+vision \\
\uquery{} &  & text+query \\
\kiwi{}  & Wan2.2 5B & query \\
\wanctrl{}  & Wan2.1 14B & text+query \\
\bottomrule
\end{tabular}
\tightcaption{Model used for diagnosis.}
\label{tab:interfaces}
\vspace{-2em} % 微调表格与下方文字的间距
\end{wraptable}
For each relation-editing example, we extract a pre-connector feature \(\pre^{(i)}\) and the corresponding post-connector DiT-facing feature \(\post^{(i)}=\conn(\pre^{(i)})\). Depending on the model, these features are pooled over hidden states, text-plus-query tokens, or learnable query tokens. All probing experiments use composition-level train/validation/test splits, so that no four-cell source scene appears in both training and testing. Results are averaged over 3 seeds.

The diagnostics below answer five questions: (1) which variables survive connector alignment, (2) whether alignment is geometrically non-trivial, (3) how query evidence is redistributed, (4) how the DiT consumes condition tokens, and (5) whether generated-video failures concentrate on the same structural axes. 

\section{Empirical Analysis of VLM-to-DiT Semantic Preservation}
\label{sec:results}

\subsection{Geometric Reconfiguration of Post-connector Representations}
\label{sec:geometry}

\noindent \textbf{Experimental setup and metrics.}
This diagnostic tests whether connector alignment merely projects VLM features into the original DiT text embedding space or substantially reconfigures their global geometry. For each model, we extract the pre-connector embeddings and their corresponding post-connector DiT embeddings from all samples to form the feature matrices $X, Y \in \mathbb{R}^{n\times d}$. For a centered feature matrix $X$ with singular values $\{\sigma_k\}$, we define the normalized singular mass $p_k=\sigma_k/\sum_j\sigma_j$ and the effective rank as $r_{\mathrm{eff}}(X)=\exp\left(-\sum_k p_k\log p_k\right)$. We also measure feature variance and linear CKA (Centered Kernel Alignment) between the centered matrices $X$ and $Y$ as follows:
\begin{equation}
    \mathrm{Var}(X)=\frac{1}{nd}\sum_{i=1}^{n}\sum_{j=1}^{d}(X_{ij}-\bar X_j)^2, \qquad
    \mathrm{CKA}(X,Y)=\frac{\|X^\top Y\|_F^2}{\|X^\top X\|_F\,\|Y^\top Y\|_F},
\end{equation}
where the effective rank measures the spread of singular mass across feature dimensions, feature variance measures representation scale, and CKA measures global similarity between the pre- and post-connector condition spaces.
\begin{table*}[t]
    \centering
    \small
    \begin{tabular}{l ccc ccc c}
        \toprule
        \multirow{2}{*}{\textbf{Model}} & \multicolumn{3}{c}{\textbf{Effective rank}} & \multicolumn{3}{c}{\textbf{Feature variance}} & \multirow{2}{*}{\textbf{CKA}} \\
        \cmidrule(lr){2-4} \cmidrule(lr){5-7}
        & Pre & Post & \(\Delta\) & Pre & Post & \(\Delta\) & \\
        \midrule
        \uhid{}    & 1096.1 & 458.2  & \(-58.2\%\) & \(3.25\!\times\!10^{-1}\) & \(6.17\!\times\!10^{-2}\) & \(-81.0\%\) & .840 \\
        \uquery{}  & 871.3  & 350.5  & \(-59.8\%\) & \(4.27\!\times\!10^{-2}\) & \(4.20\!\times\!10^{-3}\) & \(-90.2\%\) & .896 \\
        \kiwi{}    & 748.9  & 824.8  & \(+10.1\%\) & \(5.26\!\times\!10^{-4}\) & \(1.68\!\times\!10^{-5}\) & \(-96.8\%\) & .966 \\
        \wanctrl{} & 892.0  & 1041.2 & \(+16.7\%\) & \(3.77\!\times\!10^{-1}\) & \(1.00\!\times\!10^{-3}\) & \(-99.7\%\) & .871 \\
        \bottomrule
    \end{tabular}
    \tightcaption{Representation geometry before and after connector alignment. Connector alignment substantially changes scale and rank structure for every model, showing that DiT-facing condition representations are not passive projections of VLM-side features.}
    \label{tab:p1_geometry}
\end{table*}

\noindent \textbf{Results and analysis.}
Tab.~\ref{tab:p1_geometry} shows that connector alignment substantially reconfigures the condition space. \uhid{} and \uquery{} undergo large effective-rank reductions. In contrast, \kiwi{} and \wanctrl{} show an increase in effective rank after alignment, yet both exhibit a severe collapse in feature variance. 
% \kiwi{} preserves edited-slot readout despite a severe variance drop, whereas \wanctrl{} still loses target-value semantics despite increased rank. 
These findings establish that connector alignment is a profound geometric transformation rather than a mere passive format conversion. However, global geometry alone cannot determine which semantic variables survive: models with similar CKA or rank trends can differ sharply in their preservation of slot, value, and role information. This discrepancy motivates the variable-level semantic decodability in Sec.~\ref{sec:linear_readout} and the subsequent token-route analyses.

\subsection{Selective Preservation of Relational Variables under Connector Alignment}
\label{sec:linear_readout}

\noindent \textbf{Experimental setup and metrics.} This diagnostic measures whether each task-critical relational information remains linearly accessible before and after connector alignment. We cover five information types including the type of edit attribute, edited slot, target value, edited objected and reference object in this experiment. For each label family \(y\) and representation stage \(r\in\{\mathrm{pre},\mathrm{post}\}\), we train a linear classifier
\begin{equation}
    f_y^r(h)=\operatorname{softmax}(W_y^r\,\phi_r(h)+b_y^r),
\end{equation}
where \(\phi_r(\cdot)\) denotes standardized PCA features with dimension \(128\) fit on the training set. The probes are optimized by SGD with learning rate \(0.1\), weight decay \(10^{-4}\), momentum \(0.9\). We adopt accuracy as the main metric. Since target-value label spaces differ by attribute type, the probes for target attribute value for three attribute types are trained respectively, based on which the average metric is reported.

\begin{table*}[t]
\centering
\small
\setlength{\tabcolsep}{4pt}
\begin{tabular}{lccccc}
\toprule
\textbf{Variable} & \textbf{Chance} & \textbf{\uhid{}} & \textbf{\uquery{}} & \textbf{\kiwi{}} & \textbf{\wanctrl{}} \\
\midrule
Attribute type & \(1/3\) & \(.988\to.977\) & \(.998\to.998\) & \(.999\to.997\) & \(.999\to.999\) \\
Edited slot & \(1/4\) & \(.260\to.248\) & \(.923\to.739\) & \(.992\to.987\) & \(.897\to.863\) \\
Target value avg. & \(\approx.107\) & \(.220\to.214\) & \(.517\to.491\) & \(.502\to.485\) & \(.345\to.306\) \\
Edited object & \(1/15\) & \(.340\to.341\) & \(.375\to.426\) & \(.470\to.420\) & \(.462\to.471\) \\
Reference object & \(1/15\) & \(.343\to.342\) & \(.415\to.444\) & \(.586\to.560\) & \(.355\to.345\) \\
\bottomrule
\end{tabular}
\tightcaption{Pre-connector \(\to\) post-connector linear-probe accuracy on \bench{}. Target value is the unweighted average over action, material, and color target probes; its chance level is the corresponding average over the attribute-specific label spaces. Full per-attribute target results are in Appendix~\ref{app:full_results}.}
\label{tab:p0_main}
\vspace{-0.2in}
\end{table*}

\noindent \textbf{Results and analysis.}
Tab.~\ref{tab:p0_main}
% and Fig.~\ref{fig:semantic_deltas} 
demonstrates a selective preservation pattern. (1) Performance on the attribute-type task remains near-perfect after alignment across all models. Thus, connectors reliably preserve the coarse edit category, e.g., whether an instruction concerns color, material, or action. (2) Results from the edited-slot task separate the models into distinct regimes. While all four models exhibit decreased accuracy with post-connector embeddings, \uhid{} fails entirely on this task, yielding near-random performance with both pre- and post-connector features.
% \uhid{} never exposes slot semantics strongly, so its small delta should not be interpreted as successful preservation: both pre- and post-connector accuracies are close to the four-way chance level. 
\uquery{} exposes the edited slot before alignment, but the post-connector embedding loses a large fraction of that accessibility, dropping from \(.923\) to \(.739\). Other two models shows a smaller but consistent decline. (3) Target-value semantics are more fragile. Even models with strong slot prediction ability lose target-value accessibility after alignment. 
% \wanctrl{} drops across action, material, and color targets---action \(.202\to.174\), material \(.390\to.350\), and color \(.444\to.394\)---despite using the Wan backbone, which rules out a simple weak-backbone explanation. 
(4) Object-role information does not follow a uniform degradation pattern: \kiwi{} loses edited-object and reference-object decodability, while \uquery{} gains object-role accessibility after alignment. This strengthens our hypothesis that connector alignment does not simply erase all fine-grained information, but rather fractures the relational edit program by \textbf{weakening and redistributing specific semantic components}. While capable of preserving coarse edit intent, the connector fails to reliably preserve the fine-grained slot--value--role binding required by our benchmark.

\subsection{Token-Route Decomposition of Query Semantics}
\label{sec:token_route}
\noindent \textbf{Experimental setup and metrics.}
This diagnostic investigates where semantic evidence resides within models utilizing extra query tokens. Concretely, we decompose \uquery{} into three feature views: a \textbf{mixed} view containing embeddings of both text and query tokens, a \textbf{text-only} view, and a \textbf{query-only} view. We then measure edited-slot and target-value probing ability from each view using the same methodology as in Sec.~\ref{sec:linear_readout}. To localize evidence within the mixed route, we compute token-level margin contributions for the edited-slot classifier. For a token \(t\), correct class \(c^\star\), and strongest incorrect class \(\hat c\), the approximate margin contribution is
\begin{equation}
    m_{it}=\left\langle w_{c^\star}-w_{\hat c},\,\psi(h_{it})\right\rangle,
\end{equation}
where \(\psi(h_{it})\) is the token contribution after the same linear feature map. We report the total margin \(\sum_t m_{it}\), normalized positive-contribution entropy
\begin{equation}
    H_i^+=-\frac{1}{\log T}\sum_{t=1}^{T} q_{it}\log q_{it},\qquad
    q_{it}=\frac{\max(m_{it},0)}{\sum_j \max(m_{ij},0)},
\end{equation}
top-1 positive mass \(\max_t q_{it}\), and the fraction of positive mass assigned to query tokens.

\begin{table}[h]
\centering
\small
\resizebox{\linewidth}{!}{%
\begin{tabular}{lccccc}
\toprule
\textbf{View} & \textbf{Attr. type} & \textbf{Edited slot} & \textbf{Material target} & \textbf{Color target} & \textbf{Reference object} \\
\midrule
Mixed & \(0.998 \to 0.998\) & \(0.923 \to 0.739\) & \(0.619 \to 0.593\) & \(0.675 \to 0.633\) & \(0.414 \to 0.444\) \\
Text only & \(1.000 \to 0.999\) & \(0.999 \to 0.988\) & \(0.897 \to 0.829\) & \(0.924 \to 0.868\) & \(0.813 \to 0.595\) \\
Query only & \(0.999 \to 0.994\) & \(0.781 \to 0.556\) & \(0.496 \to 0.528\) & \(0.551 \to 0.512\) & \(0.385 \to 0.389\) \\
\bottomrule
\end{tabular}}
\caption{\uquery{} feature-view ablation.}
\label{tab:feature_view}
\end{table}

\noindent \textbf{Results and analysis.}
The ablation results (Table~\ref{tab:feature_view}) demonstrate that a query-based model contains multiple semantic routes with varying sensitivities to connector alignment. In \uquery{}, text tokens retain strong instruction semantics: the edited-slot accuracy remains high (\(0.999 \to 0.988\)), alongside material (\(0.897 \to 0.829\)) and color (\(0.924 \to 0.868\)) targets. Conversely, query-only features are weaker and more fragile to connector alignment, with edited-slot accuracy dropping sharply (\(0.781 \to 0.556\)). The mixed route lies between these extremes, dropping from \(0.923\) to \(0.739\) on the edited-slot task. This indicates that the connector does not universally erase slot semantics from every subview; rather, it alters how the mixed route exposes this information. 

Furthermore, token-level margins reveal this routing shift at a finer scale. In \uquery{}, the mean edited-slot margin drops from \(18.08\) to \(11.13\) after alignment. The positive evidence mass contributed by query tokens decreases from \(0.874\) to \(0.623\), while the top-1 positive mass spikes from \(0.016\) to \(0.291\). This indicates a shift toward a highly concentrated, less query-dominated evidence pattern after alignment. Cross-model comparisons reinforce this interpretation: \kiwi{} maintains a stable edited-slot margin (\(24.82 \to 27.70\)), consistent with its strong slot preservation, whereas \wanctrl{} shows only a mild edited-slot margin drop (\(16.57 \to 15.99\)) but still loses target-value decodability. Therefore, while query-token conditioning provides a pathway for relational evidence, its preservation heavily depends on whether the connector alignment maintains the accessibility of slot--value--role information along that specific route.

\subsection{Attention Routing and Semantic Consumption by the DiT}
\label{sec:condition_attention}

\noindent \textbf{Experimental setup and metrics.}
This diagnostic tests how the DiT consumes the post-connector embeddings. For each model, we focus on the attention from generated video tokens to condition tokens. Formally, for attention weight \(\alpha_{\ell,t,h,q,k}\) from generated token \(q\) to condition token \(k\) at layer \(\ell\), denoising step \(t\), and head \(h\), the group attention share for condition-token group \(g\) is
\begin{equation}
    A_g=\frac{\sum_{\ell,t,h,q}\sum_{k\in g}\alpha_{\ell,t,h,q,k}}
    {\sum_{\ell,t,h,q}\sum_{k\in\mathcal{K}}\alpha_{\ell,t,h,q,k}}.
\end{equation}
We also normalize the aggregated condition-token distribution \(\bar\alpha_k\), and report entropy \(H_A=-\sum_k \bar\alpha_k\log\bar\alpha_k\), top-\(K\) mass \(M_K=\sum_{k\in\mathrm{Top}K(\bar\alpha)}\bar\alpha_k\), and head agreement, measured as the average Jaccard similarity between the top-16 condition-token sets selected by different heads. Statistics are computed on \(20\) examples per model. \uhid{} is not included in this experiment due to its lack of query tokens.

\begin{figure*}[t]
    \centering
    \includegraphics[width=0.9\textwidth]{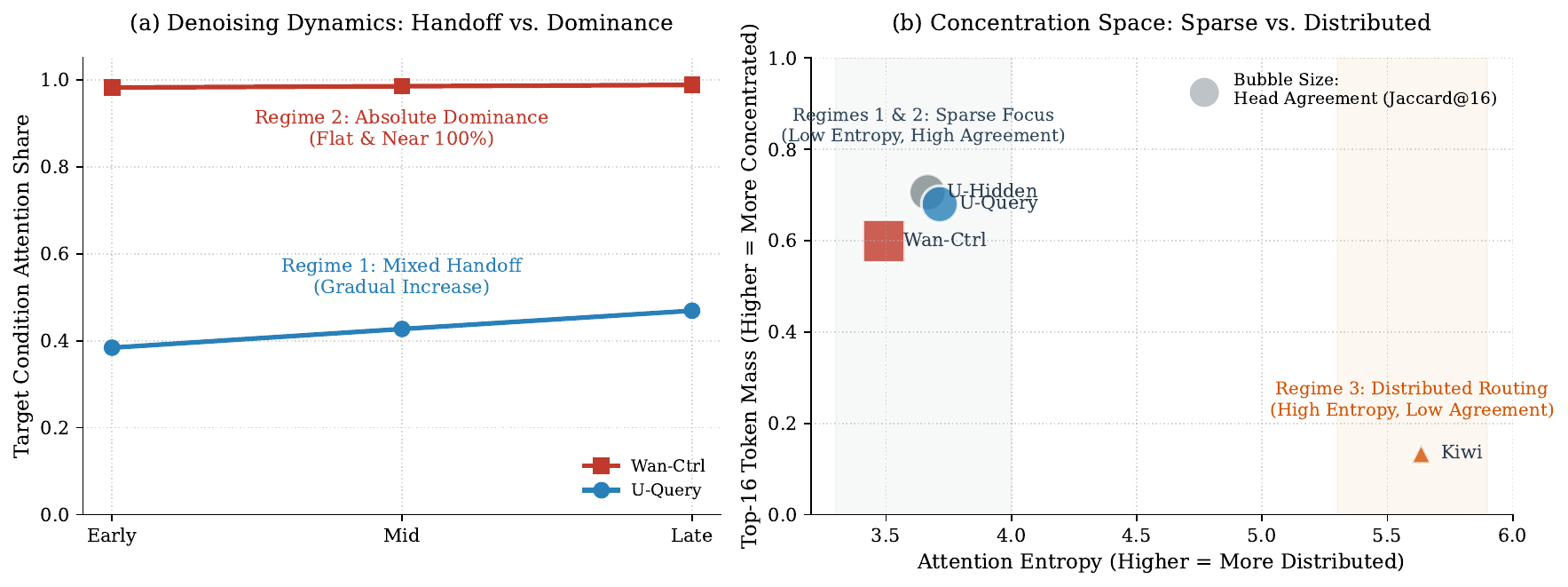}
    \tightcaption{Condition-attention routing regimes. \uquery{} gradually increases query attention while retaining text anchors; \wanctrl{} is query-dominated from the beginning; \kiwi{} distributes attention over many learnable queries with low head agreement.}
    \label{fig:attn_main}
\end{figure*}

\noindent \textbf{Results and analysis.}
Fig.~\ref{fig:attn_main} reveals three ways in which the DiT consumes post-connector embeddings. First, \uquery{} exhibits a mixed handoff regime: overall condition attention remains text-dominant (text \(.573\), query \(.427\)), while query share increases from \(.384\) early in denoising to \(.469\) late in denoising. The DiT therefore allocates increasing attention to query tokens while retaining text as a structural anchor. Second, \wanctrl{} exhibits absolute query dominance. Although it contains both text and query condition groups, almost all condition attention is assigned to query tokens: query share is \(.985\) overall and changes only slightly from early to late denoising. Its attention is also relatively concentrated, with top-32 condition tokens absorbing \(.977\) of the condition attention and a higher head top-16 Jaccard score than \uquery{}. Third, \kiwi{} uses distributed query routing: its condition group consists of learnable queries, but attention is spread across them, resulting in high entropy \((5.635)\), low top-16 mass \((.134)\), and low head agreement \((.116)\). These regimes show that attention allocation and semantic preservation are not equivalent. \wanctrl{} assigns almost all condition attention to query tokens, yet still loses target-value accessibility and exhibits structural output failures. The relevant question is therefore not only how much attention the DiT assigns to query tokens, but whether that route preserves and exposes the correct slot--value--role information.

\subsection{Output-Level Structural Failure Analysis}
\label{sec:output_audit}

\noindent \textbf{Experimental setup and metrics.}
This diagnostic connects connector-related measurements to the generated videos. Because full video inference is expensive, we evaluate a balanced subset of \(144\) examples for each model. Each generated output is assigned to one of six categories: pass (i.e. the edit is success), under-edit (i.e. the source video is not changed), wrong slot, wrong target value, wrong object/binding, or undetermined. Let \(n_{\mathrm{eval}}\) be the number of evaluated outputs, \(n_{\mathrm{pass}}\) the number of correct outputs, \(n_{\mathrm{slot}}\) the number of wrong-slot outputs, \(n_{\mathrm{bind}}\) the number of wrong-object/binding outputs, and \(n_{\mathrm{under}}\) the number of under-edited outputs. We report
\begin{equation}
    \mathrm{PassRate} = \frac{n_{\mathrm{pass}}}{n_{\mathrm{eval}}}, \qquad
    \mathrm{StructErr} = \frac{n_{\mathrm{slot}}+n_{\mathrm{bind}}}{n_{\mathrm{eval}}}, \qquad
    \mathrm{UnderEdit} = \frac{n_{\mathrm{under}}}{n_{\mathrm{eval}}}.
\end{equation}
Structural error is defined as wrong-slot plus wrong-object/binding failures. Such an experiment serves as a consistency check between representation diagnostics and output behavior.

\begin{table}[t]
\centering
\small
\begin{tabular}{lcccc}
\toprule
\multirow{2}{*}{\textbf{Model}} & \textbf{Evaluated} & \textbf{PassRate} & \textbf{StructErr} & \textbf{UnderEdit} \\
& \(n_{\mathrm{eval}}\) & Pass & Wrong slot + binding & Under-edit \\
\midrule
\uhid{}    & \(119\) & \(25.2\%\) & \(58.0\%\) & \(12.6\%\) \\
\uquery{}  & \(139\) & \(22.3\%\) & \(61.9\%\) & \(12.2\%\) \\
\kiwi{}    & \(137\) & \(38.0\%\) & \(40.9\%\) & \(14.6\%\) \\
\wanctrl{} & \(135\) & \(25.2\%\) & \(57.0\%\) & \(11.1\%\) \\
\bottomrule
\end{tabular}
\tightcaption{Output failure evaluation on \(144\) examples per model. Rates are computed among successfully evaluated outputs. Structural errors dominate under-editing across all models.}
\label{tab:output_audit}
\vspace{-0.3in}
\end{table}

\noindent \textbf{Results and analysis.}
Tab.~\ref{tab:output_audit} shows that under-editing is not the dominant failure mode. Across models, under-editing accounts for only \(11.1\%\)--\(14.6\%\) of evaluated outputs, whereas structural errors account for \(40.9\%\)--\(61.9\%\). The generated failures are therefore primarily structural rather than purely conservative: the systems often perform an edit, but fail to preserve the assignment specified by the instruction, i.e. which slot should change, which object should be edited, and which reference object or target value should govern the edit. The cross-model pattern aligns with the connector-side diagnostics without requiring a per-sample causal claim.
Specifically, \uquery{} exhibits the highest structural-error rate, consistent with its severe post-connector slot degradation. Meanwhile, \kiwi{} achieves the highest pass rate and the lowest structural-error rate, consistent with stronger slot preservation; however, it still produces numerous structural failures because target-value and role variables remain fragile. Furthermore, \wanctrl{} performs similarly to the two UniVideo variants despite using a Wan backbone rather than HunyuanVideo, supporting the conclusion that backbone replacement alone does not remove the VLM-to-DiT alignment bottleneck. 
These output-level results connect the representation diagnostics to end-task behavior: the principal failure mode is structural misassignment rather than a generic lack of editing strength.

\section{Discussion}

\noindent \textbf{Connectors are critical semantic transferers.}
The central lesson is that VLM-to-DiT alignment should not be evaluated only by end-to-end video quality. A connector may preserve coarse editing categories while failing to expose the variables that make an edit correct. One likely reason is that current training aligns VLM features to a pre-existing DiT text-conditioning space, which was not originally optimized to encode disentangled relational edit programs such as edited slots, target values, and object roles. Another possible contributor is supervision mismatch: generic T2V, I2V, and editing data may support coarse instruction following, but may not sufficiently constrain fine-grained slot--value--role preservation. In \bench{}, attribute type is almost perfectly preserved, but edited slot, target value, and object-role variables behave differently across models. This is why output metrics alone are insufficient: two systems can both understand that an instruction asks for a material edit while differing sharply in whether they edit the correct cell, transfer the correct value, or bind the correct reference object.

\noindent \textbf{Query token conditioning routes help, but introduce a routing problem.}
Query tokens are not a semantic solution by themselves. They provide a route through which text-side, visual, and relational evidence can reach the DiT, but the connector must still preserve and expose the right variables along that route. \uquery{} demonstrates the failure case: slot semantics are available before alignment but weakened in the DiT-facing mixed view. \kiwi{} demonstrates the partial success case: slot semantics can be preserved, yet target-value and role-binding semantics can still degrade. The design problem is therefore not merely whether to add queries, but how to make the query route preserve the whole slot--value--role information.

\noindent \textbf{Backbone replacement is not enough.}
Since \wanctrl{} retains a UniVideo-style VLM utilization but uses a Wan DiT backbone apart from HunyuanVideo, its persistent target-value degradation shows that a stronger DiT cannot recover structural variables that the model fails to expose. This does not mean the backbone is irrelevant, but rather means that backbone quality and the connector's ability of semantic preservation are distinct requirements for a qualified video editing model.

\noindent \textbf{What future video editors should report.}
VLM-conditioned video editors that adopt the connector alignment strategy should report connector related diagnostics for task-critical variables, not only output-level instruction-following or video quality scores. Moreover, if a model relies on query tokens, performance focusing on whether slot, value, and object-role variables remain accessible after the query connector is also necessary for comprehensive analysis of the model. These diagnostics are inexpensive compared with full video inference and can reveal structural failure modes before large-scale generation studies.

\section{Conclusion}

We introduced \bench{} as a systematic diagnostic tool for VLM-to-DiT video editing models.
Across various controlled systems, connectors preserve coarse attribute categories but degrade fine-grained structural semantics such as edited slots, target values, and object roles. These representation-level losses align with output-level structural failures. Future VLM-conditioned video editors should therefore treat connector design as a semantic-preserving problem and connector-side diagnostics can be taken into account as critical experiment metrics in addition to final generation quality.

{\small
\bibliographystyle{plainnat}
\bibliography{ref}

\begin{thebibliography}{19}
\providecommand{\natexlab}[1]{#1}
\providecommand{\url}[1]{\texttt{#1}}
\expandafter\ifx\csname urlstyle\endcsname\relax
  \providecommand{\doi}[1]{doi: #1}\else
  \providecommand{\doi}{doi: \begingroup \urlstyle{rm}\Url}\fi

\bibitem[Bai et~al.(2025)Bai, Chen, Liu, Wang, Ge, Song, Dang, Wang, Wang, Tang, et~al.]{qwen25vl2025}
Shuai Bai, Keqin Chen, Xuejing Liu, Jialin Wang, Wenbin Ge, Sibo Song, Kai Dang, Peng Wang, Shijie Wang, Jun Tang, et~al.
\newblock Qwen2.5-vl technical report.
\newblock \emph{arXiv preprint arXiv:2502.13923}, 2025.

\bibitem[Basu et~al.(2023)Basu, Saberi, Bhardwaj, Chegini, Massiceti, Sanjabi, Hu, and Feizi]{basu2023editval}
Samyadeep Basu, Mehrdad Saberi, Shweta Bhardwaj, Atoosa~Malemir Chegini, Daniela Massiceti, Maziar Sanjabi, Shell~Xu Hu, and Soheil Feizi.
\newblock Editval: Benchmarking diffusion based text-guided image editing methods.
\newblock \emph{arXiv preprint arXiv:2310.02426}, 2023.

\bibitem[Chen et~al.(2025{\natexlab{a}})Chen, Xu, Pan, Hu, Qin, Goldstein, Huang, Zhou, Xie, Savarese, Xue, Xiong, and Xu]{blip3o2025}
Jiuhai Chen, Zhiyang Xu, Xichen Pan, Yushi Hu, Can Qin, Tom Goldstein, Lifu Huang, Tianyi Zhou, Saining Xie, Silvio Savarese, Le~Xue, Caiming Xiong, and Ran Xu.
\newblock Blip3-o: A family of fully open unified multimodal models--architecture, training and dataset.
\newblock \emph{arXiv preprint arXiv:2505.09568}, 2025{\natexlab{a}}.

\bibitem[Chen et~al.(2026)Chen, He, Fu, Wan, Gai, and Ye]{vino2026}
Junyi Chen, Tong He, Zhoujie Fu, Pengfei Wan, Kun Gai, and Weicai Ye.
\newblock Vino: A unified visual generator with interleaved omnimodal context.
\newblock \emph{arXiv preprint arXiv:2601.02358}, 2026.

\bibitem[Chen et~al.(2025{\natexlab{b}})Chen, Zhang, Hu, Zeng, Xue, He, Wang, Liu, Hu, and Yan]{chen2025ivebench}
Yinan Chen, Jiangning Zhang, Teng Hu, Yuxiang Zeng, Zhucun Xue, Qingdong He, Chengjie Wang, Yong Liu, Xiaobin Hu, and Shuicheng Yan.
\newblock Ivebench: Modern benchmark suite for instruction-guided video editing assessment.
\newblock \emph{arXiv preprint arXiv:2510.11647}, 2025{\natexlab{b}}.

\bibitem[HaCohen et~al.(2026)HaCohen, Brazowski, Chiprut, Bitterman, Kvochko, Berkowitz, Shalem, Lifschitz, Moshe, Porat, et~al.]{hacohen2026ltx}
Yoav HaCohen, Benny Brazowski, Nisan Chiprut, Yaki Bitterman, Andrew Kvochko, Avishai Berkowitz, Daniel Shalem, Daphna Lifschitz, Dudu Moshe, Eitan Porat, et~al.
\newblock Ltx-2: Efficient joint audio-visual foundation model.
\newblock \emph{arXiv preprint arXiv:2601.03233}, 2026.

\bibitem[Huang et~al.(2025)Huang, Wang, Yang, Lu, Yuan, Han, Hou, Zhang, Hong, Zhao, and Xu]{illumeplus2025}
Runhui Huang, Chunwei Wang, Junwei Yang, Guansong Lu, Yunlong Yuan, Jianhua Han, Lu~Hou, Wei Zhang, Lanqing Hong, Hengshuang Zhao, and Hang Xu.
\newblock Illume+: Illuminating unified mllm with dual visual tokenization and diffusion refinement.
\newblock \emph{arXiv preprint arXiv:2504.01934}, 2025.

\bibitem[Huang et~al.(2024)Huang, He, Yu, Zhang, Si, Jiang, Zhang, Wu, Jin, Chanpaisit, et~al.]{huang2024vbench}
Ziqi Huang, Yinan He, Jiashuo Yu, Fan Zhang, Chenyang Si, Yuming Jiang, Yuanhan Zhang, Tianxing Wu, Qingyang Jin, Nattapol Chanpaisit, et~al.
\newblock Vbench: Comprehensive benchmark suite for video generative models.
\newblock In \emph{Proceedings of the IEEE/CVF Conference on Computer Vision and Pattern Recognition}, pages 21807--21818, 2024.

\bibitem[Ju et~al.(2025)Ju, Wang, Zhou, Zhang, Liu, Zhao, Zhang, Li, Cai, Liu, et~al.]{ju2025editverse}
Xuan Ju, Tianyu Wang, Yuqian Zhou, He~Zhang, Qing Liu, Nanxuan Zhao, Zhifei Zhang, Yijun Li, Yuanhao Cai, Shaoteng Liu, et~al.
\newblock Editverse: Unifying image and video editing and generation with in-context learning.
\newblock \emph{arXiv preprint arXiv:2509.20360}, 2025.

\bibitem[Kong et~al.(2024)Kong, Tian, Zhang, Min, Dai, Zhou, Xiong, Li, Wu, Zhang, et~al.]{hunyuanvideo2024}
Weijie Kong, Qi~Tian, Zijian Zhang, Rox Min, Zuozhuo Dai, Jin Zhou, Jiangfeng Xiong, Xin Li, Bo~Wu, Jianwei Zhang, et~al.
\newblock Hunyuanvideo: A systematic framework for large video generative models.
\newblock \emph{arXiv preprint arXiv:2412.03603}, 2024.

\bibitem[Lin et~al.(2026)Lin, Liang, Zeng, Bai, Chen, and Shou]{kiwiedit2026}
Yiqi Lin, Guoqiang Liang, Ziyun Zeng, Zechen Bai, Yanzhe Chen, and Mike~Zheng Shou.
\newblock Kiwi-edit: Versatile video editing via instruction and reference guidance.
\newblock \emph{arXiv preprint arXiv:2603.02175}, 2026.

\bibitem[Liu et~al.(2024)Liu, Cun, Liu, Wang, Zhang, Chen, Liu, Zeng, Chan, and Shan]{liu2024evalcrafter}
Yaofang Liu, Xiaodong Cun, Xuebo Liu, Xintao Wang, Yong Zhang, Haoxin Chen, Yang Liu, Tieyong Zeng, Raymond Chan, and Ying Shan.
\newblock Evalcrafter: Benchmarking and evaluating large video generation models.
\newblock In \emph{Proceedings of the IEEE/CVF conference on computer vision and pattern recognition}, pages 22139--22149, 2024.

\bibitem[Mou et~al.(2025)Mou, Sun, Wu, Zhang, Li, Ye, Zhao, and He]{mou2025instructxunifiedvisualediting}
Chong Mou, Qichao Sun, Yanze Wu, Pengze Zhang, Xinghui Li, Fulong Ye, Songtao Zhao, and Qian He.
\newblock Instructx: Towards unified visual editing with mllm guidance.
\newblock \emph{https://arxiv.org/abs/2510.08485}, 2025.

\bibitem[Pan et~al.(2026)Pan, Tian, Zhang, Kong, Xiong, Long, Zhang, Qiu, Wang, Lv, Wu, Bo, Tang, and Zhong]{pan2026omniweavingunifiedvideogeneration}
Kaihang Pan, Qi~Tian, Jianwei Zhang, Weijie Kong, Jiangfeng Xiong, Yanxin Long, Shixue Zhang, Haiyi Qiu, Tan Wang, Zheqi Lv, Yue Wu, Liefeng Bo, Siliang Tang, and Zhao Zhong.
\newblock Omniweaving: Towards unified video generation with free-form composition and reasoning.
\newblock \emph{https://arxiv.org/abs/2603.24458}, 2026.

\bibitem[Pan et~al.(2025)Pan, Shukla, Singh, Zhao, Mishra, Wang, Xu, Chen, Li, Juefei-Xu, Hou, and Xie]{metaqueries2025}
Xichen Pan, Satya~Narayan Shukla, Aashu Singh, Zhuokai Zhao, Shlok~Kumar Mishra, Jialiang Wang, Zhiyang Xu, Jiuhai Chen, Kunpeng Li, Felix Juefei-Xu, Ji~Hou, and Saining Xie.
\newblock Transfer between modalities with metaqueries.
\newblock \emph{arXiv preprint arXiv:2504.06256}, 2025.

\bibitem[Song et~al.(2025)Song, Dong, Wang, Zhang, Xue, Yuan, Yang, Feng, Zhou, Xiao, and Wang]{querykontext2025}
Yuxin Song, Wenkai Dong, Shizun Wang, Qi~Zhang, Song Xue, Tao Yuan, Hu~Yang, Haocheng Feng, Hang Zhou, Xinyan Xiao, and Jingdong Wang.
\newblock Query-kontext: An unified multimodal model for image generation and editing.
\newblock \emph{arXiv preprint arXiv:2509.26641}, 2025.

\bibitem[{Wan Team}(2025)]{wan2025}
{Wan Team}.
\newblock Wan: Open and advanced large-scale video generative models.
\newblock \emph{arXiv preprint arXiv:2503.20314}, 2025.

\bibitem[Wei et~al.(2026)Wei, Liu, Ye, Wang, Wang, Wan, Gai, and Chen]{univideo2026}
Cong Wei, Quande Liu, Zixuan Ye, Qiulin Wang, Xintao Wang, Pengfei Wan, Kun Gai, and Wenhu Chen.
\newblock Univideo: Unified understanding, generation, and editing for videos.
\newblock \emph{arXiv preprint arXiv:2510.08377}, 2026.

\bibitem[Xiao et~al.(2024)Xiao, Wang, Zhou, Yuan, Xing, Yan, Li, Wang, Huang, and Liu]{omnigen2024}
Shitao Xiao, Yueze Wang, Junjie Zhou, Huaying Yuan, Xingrun Xing, Ruiran Yan, Chaofan Li, Shuting Wang, Tiejun Huang, and Zheng Liu.
\newblock Omnigen: Unified image generation.
\newblock \emph{arXiv preprint arXiv:2409.11340}, 2024.

\end{thebibliography}
}

\newpage
\appendix

\section{Limitations}
\label{app:limitations}
Our analysis focuses on connector-based VLM-to-DiT video editing models, where a pre-trained or independently trained VLM representation is aligned with the conditioning space of a video DiT through an explicit connector. This setting covers the design pattern adopted by the evaluated systems and directly matches our goal of diagnosing whether fine-grained relational semantics survive the connector. However, our experiments do not cover architectures that are natively trained with VLM-based conditioning throughout the video generator, or systems in which multimodal reasoning and denoising are jointly optimized without an explicit post-hoc VLM-to-DiT alignment strategy, since such models are currently not accessible. Moreover, our proposed TRACE-Edit deliberately sacrifices natural-scene diversity for causal control over slot, value, and role variables. The video grid design is not intended to approximate all real editing scenarios but rather a stress test for relational binding under known ground truth.

\section{Additional Benchmark and Protocol Details}
\label{app:details}

\paragraph{TRACE-Edit composition.}
Each \bench{} composite video is assembled from four verified atomic videos. A composite instance uses one attribute type and four different attribute values whenever possible. We draw an edited slot and a reference slot, store the object identity and attribute value for each cell, and generate two relation directions when applicable. This yields $5{,}524$ composite videos and $11{,}048$ relation-editing examples.

\paragraph{Probe configuration.}
Unless otherwise stated, linear-probe analysis probes use PCA dimension $128$, SGD with learning rate $0.1$, weight decay $10^{-4}$, momentum $0.9$, patience $30$, and train/validation/test ratio $0.7/0.15/0.15$. The reported mean and standard deviation are over split seeds $0$, $1$, and $42$.

\paragraph{Computational resources.}
All experiments were run on 8 H100 GPUs.

\section{\bench{} Construction Details}
\label{app:trace_edit_details}

This section provides the concrete construction details omitted from the main text. The implementation uses structured atomic scene specifications, generates one video per atomic specification, verifies the generated video with a VLM, and then composes only verified atomic videos into relation scenes.

\paragraph{Object and attribute-value pools.}
The current benchmark instantiation samples objects and attribute values from the fixed pools in Tab.~\ref{tab:trace_edit_pools}. Each atomic video contains one object, one attribute type, and one value from the corresponding pool.

\begin{table}[h]
\centering
\small
\resizebox{\linewidth}{!}{%
\begin{tabular}{lp{0.74\linewidth}}
\toprule
\textbf{Pool} & \textbf{Values} \\
\midrule
Objects & cup, vase, chair, table lamp, backpack, box, bottle, shoe, watch, teapot, sculpture, toy car, pen holder, picture frame, headphones \\
Color values & deep red, blue, dark green, yellow, warm orange, white, black, silver gray, teal \\
Material values & brushed metal, natural wood, transparent glass, leather, ceramic, canvas, rubber, stone, woven bamboo, wool \\
Action values & static, moving left to right, moving right to left, falling from top to bottom, floating upward from bottom to top, moving from upper-left to lower-right, moving from lower-right to upper-left, moving from lower-left to upper-right, moving from upper-right to lower-left \\
\bottomrule
\end{tabular}}
\caption{Object and attribute-value pools used by \bench{}. The target-value probe is evaluated inside each attribute family because the color, material, and action value spaces are disjoint.}
\label{tab:trace_edit_pools}
\end{table}

\paragraph{Atomic prompt construction and video generation.}
For an atomic specification, the object description is built compositionally. Color and material values are rendered as appearance modifiers, while action values are rendered as motion phrases. For example, the structured tuple \((o=\mathrm{vase}, a=\mathrm{material}, v=\mathrm{transparent\ glass})\) yields an object description such as ``a transparent-glass vase,'' and \((o=\mathrm{toy\ car}, a=\mathrm{action}, v=\mathrm{moving\ left\ to\ right})\) yields ``a toy car moving left to right.'' The description is then wrapped by one of the following templates:
\begin{quote}
\small
The center of the frame contains only one \{object description\}; the subject is clear, the background is simple, and the camera remains stable.\\
One \{object description\} is located at the center of the frame; the background is clean and the overall motion is natural and stable.\\
The video contains only one \{object description\}; the subject is complete and clear, and the scene is simple and stable.
\end{quote}
Each accepted atomic description is generated as an 81-frame clip at \(480\times832\) resolution with the text-to-video generator described in the main paper.

\paragraph{VLM filtering prompt.}
Before composition, each atomic video is verified by a VLM using a structured checklist. The verifier receives the sampled video frames and a prompt of the following form:
\begin{quote}
\small
Please carefully watch this video and answer each question below. For each item, answer ``yes'' or ``no'' and briefly explain the reason. Finally, provide an overall judgment.\\[2pt]
1. Does the video contain a unique central subject \{object\}?\\
2. Is the \{attribute label\} of \{object\} equal to \{value\}?\\[2pt]
Please strictly output JSON only:
\texttt{\{"checks": [\{"id": 1, "question": "...", "answer": "yes/no", "reason": "..."\}, ...], "all\_pass": true/false\}}.
\end{quote}
Only atomic videos for which the parsed field \texttt{all\_pass} is true are admitted to the verified pool; failed, missing, or unparsable verifier outputs are excluded. The verifier script can also be applied to composed grids by expanding the same checklist over the ordered slots: for each slot \(s\in\{\mathrm{top\mbox{-}left},\mathrm{top\mbox{-}right},\mathrm{bottom\mbox{-}left},\mathrm{bottom\mbox{-}right}\}\), it asks whether the expected object exists in that slot and whether its attribute value matches the scene specification. In our implementation the verifier uses deterministic decoding, samples 9 frames from each video, and parses the returned JSON decision.

\paragraph{Grid composition.}
Verified atomic videos are grouped by attribute type and value. For each grid scene, we sample one attribute type, select four distinct values from that type, choose one verified atomic video for each value, and randomly assign the four videos to the ordered slots
\[
S=(\mathrm{top\mbox{-}left},\mathrm{top\mbox{-}right},\mathrm{bottom\mbox{-}left},\mathrm{bottom\mbox{-}right}).
\]
The four clips are synchronized to their common usable length, resized into equal tiles, and pasted frame-by-frame into a \(2\times2\) canvas. The grid record stores the ordered slots, the atomic video identifier and path for each cell, each cell's object metadata, the selected edited slot \(s_e\), the selected reference slot \(s_r\), and a \texttt{split\_group\_id} derived from the unordered set of four atomic video identifiers. This group id is used for composition-level train/validation/test splits, preventing the same four-cell visual scene from appearing in multiple partitions.

\paragraph{Relation-instruction templates and labels.}
Each composed scene contributes two directed relation examples: the sampled direction \((s_e\to s_r)\) and the inverted direction \((s_r\to s_e)\). The two examples share the same video but have different edited slots, reference slots, target values, and object-role labels. The instruction templates are:
\begin{quote}
\small
\textbf{Color/material:} Change the color/material of the object at \{edited slot\} to match the object at \{reference slot\}. After editing, what is the color/material of the object at \{edited slot\}?\\
\textbf{Action:} Make the motion trajectory of the object at \{edited slot\} consistent with the object at \{reference slot\}. After editing, what is the motion trajectory of the object at \{edited slot\}?
\end{quote}
For each relation row, the stored labels include \texttt{attribute\_type}, \texttt{source\_value}, \texttt{target\_value}, \texttt{edited\_slot}, \texttt{reference\_slot}, \texttt{edited\_object\_name}, \texttt{reference\_object\_name}, \texttt{direction\_tag}, and the full \texttt{scene\_spec}. This metadata is what enables the connector-side probes for slot, target value, and object-role variables.

\paragraph{Dataset scale after construction.}
The final dataset contains \(5{,}524\) composite videos and \(11{,}048\) directed relation examples. Since every grid contributes one forward and one inverted instruction, the relation-direction split is exactly \(5{,}524/5{,}524\). The attribute distribution is \(6{,}490\) color, \(3{,}096\) material, and \(1{,}462\) action examples.

\section{VLM Judge Prompt for Output-Level Failure Analysis}
\label{app:vlm_judge_prompt}

This section gives the VLM-as-judge template used for the output-level structural failure analysis in Sec. 5.5. The evaluator receives two videos in a fixed order: the original video before editing and the generated video after editing. We use a two-stage prompt to avoid conflating ``no visible edit'' with wrong-slot or wrong-binding edits. The implementation uses the Qwen/Qwen3-VL-235B-A22B-Thinking-FP8 judge by default, deterministic decoding with temperature 0, a 4096-token generation limit, samples 9 frames per video, and requests JSON-only outputs. Template variables such as \texttt{\{instruction\}}, \texttt{\{edited\_side\}}, and \texttt{\{target\_value\}} are filled from the relation metadata.

\subsection{Stage 1: edit-activation prompt}

The first prompt only asks whether the generated video contains a sufficiently visible edit relative to the input. It intentionally does not judge whether the edit is in the correct slot or has the correct target value.

\begin{lstlisting}[style=promptstyle]
You will see two videos in the following fixed order:
- Video 1: the original video before editing.
- Video 2: the edited/generated result.

Compare the two videos and, using the provided original-scene information and target edit request, judge only whether a sufficiently visible edit has occurred.

Task background:
- This is a video-editing result verification task.
- In this stage, judge only whether Video 2 contains an obvious, readable, and edit-related change compared with Video 1.
- Do not require the visible change to be in the correct slot, on the correct object, or with the correct target value.
- If a clear change occurs, even if the slot, object, binding, or target value is wrong, output edit_activation_sufficient = true.
- Output edit_activation_sufficient = false only when the overall change is still insufficient. In that case, choose an activation_failure_type.

Original scene information:
{scene_text}

Target edit information:
- instruction: {instruction}
- layout_type: {layout}
- attribute_type: {attribute_type}
- edited_side: {edited_side} ({edited_position})
- reference_side: {reference_side} ({reference_position})
- edited_object_name: {edited_object_name}
- reference_object_name: {reference_object_name}
- source_value: {source_value}
- target_value: {target_value}

If edit_activation_sufficient = false, activation_failure_type must be one of:
1. no_visible_change: Video 2 is almost unchanged from Video 1; the target object or target region barely changes.
2. partial_or_non_target_change: some change is visible, but it is mainly weak, local, on a non-target attribute, on a non-target object, or not sufficient to count as real edit activation.
3. object_missing_or_unreadable: after editing, the target region/object disappears, becomes severely blurred, is covered by the background, or cannot be read.

Output rules:
- If edit_activation_sufficient = true, activation_failure_type must be null.
- If edit_activation_sufficient = false, activation_failure_type must be one of the three categories above.
- If the case is genuinely impossible to judge, output null for both fields.

Strictly output JSON only:
{
  "edit_activation_sufficient": true/false/null,
  "activation_failure_type": "no_visible_change/partial_or_non_target_change/object_missing_or_unreadable/null",
  "confidence": "high/medium/low",
  "reason_brief": "one sentence explaining whether Video 2 shows a visible edit relative to Video 1 and, if insufficient, which activation-failure type applies"
}
\end{lstlisting}

\subsection{Stage 2: structural failure-detail prompt}

The second prompt is applied to outputs that are not rejected by the activation stage. It asks whether the visible edit is assigned to the correct slot, object, reference binding, and target value.

\begin{lstlisting}[style=promptstyle]
You will see two videos in the following fixed order:
- Video 1: the original video before editing.
- Video 2: the edited/generated result.

Compare the two videos and, using the provided original-scene information and target edit request, judge whether the main edit location is correct and whether the target edit is sufficiently clear to determine the final attribute value.

Task background:
- The previous stage only checked whether Video 2 likely contains a visible edit relative to Video 1.
- This stage checks whether the visible change mainly occurs in the correct slot, on the correct edited object, and under the correct reference binding.
- A video may contain a clear change while still editing the wrong slot or object, using the wrong reference, or changing the target too weakly to read the final attribute value.
- Therefore, also judge targeted_edit_sufficient. Only when the edit on the correct target object is clear, sufficient, and readable should target_correct be judged.
- If targeted_edit_sufficient = false, target_correct must be null.
- If the main problem is slot_correct = false, edited_object_correct = false, or reference_binding_correct = false, then targeted_edit_sufficient and target_correct should preferably be null instead of forced.
- If the evidence is insufficient for a reliable decision, output null.

Original scene information:
{scene_text}

Target edit information:
- instruction: {instruction}
- layout_type: {layout}
- attribute_type: {attribute_type}
- edited_side: {edited_side} ({edited_position})
- reference_side: {reference_side} ({reference_position})
- edited_object_name: {edited_object_name}
- reference_object_name: {reference_object_name}
- source_value: {source_value}
- target_value: {target_value}

Answer the following five structured questions:
1. slot_correct: relative to Video 1, does the main edit change in Video 2 occur at the correct edited_side?
2. edited_object_correct: is the object that mainly changed the expected edited_object_name?
3. reference_binding_correct: is the reference object and reference relation understood correctly, without confusing the reference object or binding?
4. targeted_edit_sufficient: if the change is in the correct slot and on the target object, is the edit clear, sufficient, and stable enough to reliably judge the final attribute value? If the change is too weak, too local, occluded, or unreadable, output false. If the previous fields already indicate a wrong slot, object, or binding, output null.
5. target_correct: only judge this when targeted_edit_sufficient = true. In that case, has the target object in the correct slot changed to target_value for the requested attribute_type? If targeted_edit_sufficient is not true, output null.

Strictly output JSON only:
{
  "slot_correct": true/false/null,
  "edited_object_correct": true/false/null,
  "reference_binding_correct": true/false/null,
  "targeted_edit_sufficient": true/false/null,
  "target_correct": true/false/null,
  "confidence": "high/medium/low",
  "reason_brief": "one sentence explaining whether the edit is correct or whether it fails on slot/object/binding/target sufficiency/target value"
}
\end{lstlisting}

\subsection{Failure-label derivation}

The structured fields above are normalized before computing the primary failure label. Let \texttt{activation\_failure\_type} be the Stage-1 subtype. The priority order is:

\begin{lstlisting}[style=promptstyle]
if edit_activation_sufficient is false:
    primary_failure = activation_failure_type if available else "under_edit"
elif slot_correct is false:
    primary_failure = "wrong_slot"
elif edited_object_correct is false or reference_binding_correct is false:
    primary_failure = "wrong_object_or_binding"
elif targeted_edit_sufficient is false or under_edit is true:
    primary_failure = "under_edit"
elif target_correct is false:
    primary_failure = "wrong_target_value"
elif all required fields are valid and indicate success:
    primary_failure = "pass"
else:
    primary_failure = "undetermined"
\end{lstlisting}

This two-stage design is why Sec.~\ref{sec:output_audit} separates conservative under-editing from structural misassignment: visible but misrouted edits are not counted as under-editing merely because they fail the final task.

\section{Nonlinear Probing Experiment}
\label{app:nonlinear-probe-control}

This appendix tests whether the semantic degradation observed in the linear-probe analysis of Sec.~\ref{sec:linear_readout} is merely an artifact of linear separability. We repeat the same probing protocol with a higher-capacity, two-layer MLP probe. The goal is not to replace the linear probe, but to provide a capacity-control diagnostic.

\paragraph{Experimental setup and metrics.}
All feature extraction, task labels, composition-level train/validation/test splits, and train-only PCA preprocessing are kept identical to the linear-probe analysis protocol in Sec.~\ref{sec:linear_readout} and Appendix~\ref{app:details}. In particular, we use PCA dimension 128, group splits with ratio $0.7/0.15/0.15$, and split seeds $\{0,1,42\}$. The only difference is replacing the linear classifier with a two-layer MLP:
\begin{equation}
    f^r_y(h) = \mathrm{softmax}\left(W_2 \, \mathrm{Dropout}_{0.2}\left(\mathrm{GELU}(W_1 \phi_r(h) + b_1)\right) + b_2\right),
\end{equation}
where $\phi_r(\cdot)$ denotes the same standardized PCA feature used by the linear probe, the hidden width is 256, and the output dimension is the number of classes for label family $y$. The MLP is trained with AdamW, learning rate $10^{-3}$, weight decay $10^{-4}$, batch size 512, maximum 300 epochs, and early stopping patience 30. Accuracy is reported as mean $\pm$ standard deviation over the three split seeds. As in the main paper, the target-value average is the unweighted mean over action, material, and color target probes.

\begin{table}[htbp]
\centering
\small
\setlength{\tabcolsep}{5pt}
\begin{tabular}{lcccccc}
\toprule
Scope & Linear pre & Linear post & Linear gap & MLP pre & MLP post & MLP gap \\
\midrule
All subtasks & .541 & .523 & -.018 & .559 & .552 & -.007 \\
Excluding attribute type & .465 & .445 & -.021 & .485 & .477 & -.008 \\
\bottomrule
\end{tabular}
\caption{Global comparison between linear and two-layer MLP probes. Values are mean accuracies over the four model cases, seven subtasks, and three split seeds. The gap is post-connector minus pre-connector accuracy.}
\label{tab:mlp-global-summary}
\end{table}

\begin{table*}[htbp]
\centering
\small
\resizebox{\linewidth}{!}{%
\begin{tabular}{lccccc}
\toprule
\textbf{Variable} & \textbf{Chance} & \textbf{\uhid{}} & \textbf{\uquery{}} & \textbf{\kiwi{}} & \textbf{\wanctrl{}} \\
\midrule
Attribute type & $1/3$ & $.999{\pm}.001 \rightarrow .996{\pm}.001$ & $1.000{\pm}.001 \rightarrow 1.000{\pm}.001$ & $1.000{\pm}.000 \rightarrow 1.000{\pm}.000$ & $.999{\pm}.000 \rightarrow 1.000{\pm}.000$ \\
Edited slot & $1/4$ & $.248{\pm}.011 \rightarrow .252{\pm}.003$ & $.952{\pm}.002 \rightarrow .812{\pm}.013$ & $.998{\pm}.002 \rightarrow .994{\pm}.003$ & $.938{\pm}.008 \rightarrow .921{\pm}.002$ \\
Target value avg. & $\approx .107$ & $.217{\pm}.013 \rightarrow .236{\pm}.013$ & $.534{\pm}.009 \rightarrow .517{\pm}.003$ & $.527{\pm}.010 \rightarrow .497{\pm}.005$ & $.360{\pm}.005 \rightarrow .353{\pm}.011$ \\
Edited object & $1/15$ & $.337{\pm}.013 \rightarrow .335{\pm}.017$ & $.465{\pm}.001 \rightarrow .488{\pm}.015$ & $.568{\pm}.006 \rightarrow .545{\pm}.002$ & $.481{\pm}.015 \rightarrow .499{\pm}.007$ \\
Reference object & $1/15$ & $.341{\pm}.001 \rightarrow .331{\pm}.011$ & $.414{\pm}.015 \rightarrow .489{\pm}.015$ & $.625{\pm}.010 \rightarrow .612{\pm}.004$ & $.362{\pm}.012 \rightarrow .371{\pm}.008$ \\
\bottomrule
\end{tabular}}
\caption{Pre-connector $\rightarrow$ post-connector two-layer MLP probe accuracy on TRACE-EDIT. The MLP improves absolute accuracy compared with the linear probe, but the main post-connector degradation patterns remain visible.}
\label{tab:mlp-readout-main}
\end{table*}

\begin{table*}[t]
\centering
\small
\resizebox{\linewidth}{!}{%
\begin{tabular}{lccccc}
\toprule
Target-value probe & \uhid{} & \uquery{} & \kiwi{} & \wanctrl{} \\
\midrule
Action target & $.181{\pm}.014 \rightarrow .222{\pm}.030$ & $.279{\pm}.013 \rightarrow .292{\pm}.022$ & $.319{\pm}.018 \rightarrow .274{\pm}.011$ & $.198{\pm}.031 \rightarrow .222{\pm}.030$ \\
Material target & $.242{\pm}.013 \rightarrow .257{\pm}.006$ & $.634{\pm}.024 \rightarrow .620{\pm}.021$ & $.672{\pm}.021 \rightarrow .649{\pm}.023$ & $.404{\pm}.009 \rightarrow .402{\pm}.005$ \\
Color target & $.229{\pm}.014 \rightarrow .228{\pm}.009$ & $.687{\pm}.012 \rightarrow .640{\pm}.007$ & $.590{\pm}.010 \rightarrow .567{\pm}.017$ & $.479{\pm}.027 \rightarrow .434{\pm}.010$ \\
\bottomrule
\end{tabular}}
\caption{Full target-value MLP probe accuracy, pre-connector $\rightarrow$ post-connector. Chance levels are $1/9$ for action, $1/10$ for material, and $1/9$ for color.}
\label{tab:mlp-target-breakdown}
\end{table*}

\paragraph{Results and analysis.}
Tab.~\ref{tab:mlp-global-summary} shows that the two-layer MLP increases absolute accuracy for both pre- and post-connector representations. The post-connector side benefits slightly more, so the average connector gap shrinks from $-.018$ to $-.007$ over all subtasks, and from $-.021$ to $-.008$ after excluding the nearly saturated attribute-type task. This indicates that some residual information is nonlinearly accessible after connector alignment. However, the average gap does not become a stable positive effect, so the main linear-probe conclusions are not explained away by limited probing model capacity.

The variable-level results in Tab.~\ref{tab:mlp-readout-main} further support this interpretation. Attribute type remains saturated for every model case, confirming that both linear and nonlinear probes easily recover the coarse edit category. Results on edited-slot task still shows the same structural pattern as the linear probe: \uquery{} retains a large post-connector drop ($.952 \rightarrow .812$), \wanctrl{} shows a smaller but consistent drop ($.938 \rightarrow .921$), \kiwi{} remains near saturation with only a small gap, and \uhid{} stays near the chance level. Thus,\textbf{ the key slot-degradation finding is not an artifact of using a linear classifier}.

Target-value results remain more fragile than attribute-type or slot tasks. As shown in Tab.~\ref{tab:mlp-target-breakdown}, color targets still degrade after connector alignment for \uquery{}, \kiwi{}, and \wanctrl{}; material targets also remain lower post-connector for \uquery{} and \kiwi{}, and nearly unchanged for \wanctrl{}. Action targets are weaker and more capacity-sensitive, with small positive gaps for \uhid{}, \uquery{}, and \wanctrl{} under the MLP, so we do not treat their sign as a strong conclusion. Rather than that, nonlinear-probe analysis recovers part of the absolute target-value signal, but does not remove the post-connector weakness for the higher-accuracy color and material target probes.

Object-role variables also remain consistent with the linear-probe analysis. \uquery{} again shows higher post-connector accessibility for edited and reference object names, suggesting that connector alignment can redistribute object information rather than simply erase it. In contrast, \kiwi{} still decreases on both object-role probes despite the MLP's higher absolute accuracy. These patterns reinforce the main paper's claim that connector alignment selectively reshapes the relational edit program: it preserves coarse intent, leaves some nonlinear residual signal, but still weakens or redistributes the slot--value--role information required for correct editing.

\paragraph{Takeaway.}
The MLP probe should be interpreted as a capacity-control and residual-information diagnostic. It shows that post-connector representations contain additional nonlinearly decodable information, but it does not reverse the main conclusions from the linear probe. Fine-grained structural variables, especially edited slot and target value, remain less reliably accessible after connector alignment than the coarse attribute type.

\section{Full Semantic Decodability Results}
\label{app:full_results}

Tab.~\ref{tab:target_values_full} expands the target-value column in Tab.~\ref{tab:p0_main}. Target values are evaluated separately inside each attribute type because action, material, and color have different label spaces.

\begin{table}[h]
\centering
\small
\resizebox{\linewidth}{!}{%
\begin{tabular}{lccc}
\toprule
\textbf{Interface} & \textbf{Action target} & \textbf{Material target} & \textbf{Color target} \\
\midrule
\uhid{} & $0.197{\pm}0.012 \to 0.196{\pm}0.031$ & $0.242{\pm}0.014 \to 0.239{\pm}0.016$ & $0.221{\pm}0.016 \to 0.208{\pm}0.011$ \\
\uquery{} & $0.258{\pm}0.027 \to 0.248{\pm}0.007$ & $0.619{\pm}0.014 \to 0.593{\pm}0.009$ & $0.675{\pm}0.013 \to 0.633{\pm}0.012$ \\
\kiwi{} & $0.300{\pm}0.024 \to 0.294{\pm}0.018$ & $0.623{\pm}0.016 \to 0.595{\pm}0.032$ & $0.583{\pm}0.009 \to 0.566{\pm}0.021$ \\
\wanctrl{} & $0.202{\pm}0.014 \to 0.174{\pm}0.014$ & $0.390{\pm}0.019 \to 0.350{\pm}0.009$ & $0.444{\pm}0.008 \to 0.394{\pm}0.028$ \\
\bottomrule
\end{tabular}}
\caption{Target-value probe accuracy, pre-connector $\to$ post-connector.}
\label{tab:target_values_full}
\end{table}

\section{Feature-View Ablation for \uquery{}}
\label{app:feature_view}

Tab.~\ref{tab:feature_view_detail} shows that text tokens retain strong instruction semantics, while query-only features are weaker and more vulnerable to connector-side slot loss. These are feature-view ablations on the same full extracted features, not separate input-mode runs.

\begin{table}[h]
\centering
\small
\resizebox{\linewidth}{!}{%
\begin{tabular}{lccccc}
\toprule
\textbf{View} & \textbf{Attr. type} & \textbf{Edited slot} & \textbf{Material target} & \textbf{Color target} & \textbf{Reference object} \\
\midrule
Mixed & $0.998{\pm}0.002 \to 0.998{\pm}0.003$ & $0.923{\pm}0.007 \to 0.739{\pm}0.031$ & $0.619{\pm}0.014 \to 0.593{\pm}0.009$ & $0.675{\pm}0.013 \to 0.633{\pm}0.012$ & $0.414{\pm}0.001 \to 0.444{\pm}0.004$ \\
Text only & $1.000{\pm}0.000 \to 0.999{\pm}0.000$ & $0.999{\pm}0.001 \to 0.988{\pm}0.002$ & $0.897{\pm}0.013 \to 0.829{\pm}0.011$ & $0.924{\pm}0.005 \to 0.868{\pm}0.014$ & $0.813{\pm}0.012 \to 0.595{\pm}0.056$ \\
Query only & $0.999{\pm}0.001 \to 0.994{\pm}0.003$ & $0.781{\pm}0.016 \to 0.556{\pm}0.031$ & $0.496{\pm}0.017 \to 0.528{\pm}0.016$ & $0.551{\pm}0.009 \to 0.512{\pm}0.001$ & $0.385{\pm}0.014 \to 0.389{\pm}0.002$ \\
\bottomrule
\end{tabular}}
\caption{\uquery{} feature-view ablation.}
\label{tab:feature_view_detail}
\end{table}

\section{Token-Level Margin Contribution}
\label{app:token_contrib}

Token contribution analysis is run on the edited-slot diagnostic with PCA dimension $256$. The table reports mean decision margin, normalized contribution entropy, top-1 positive contribution mass, and the fraction of positive contribution mass assigned to query tokens when query tokens exist.

\begin{table}[h]
\centering
\small
\begin{tabular}{lcccc}
\toprule
\textbf{Interface} & \textbf{Margin} & \textbf{Norm. entropy} & \textbf{Top-1 mass} & \textbf{Query positive mass} \\
\midrule
\uhid{} & $-1.14 \to -2.09$ & $0.957 \to 0.917$ & $0.004 \to 0.068$ & -- \\
\uquery{} & $18.08 \to 11.13$ & $0.954 \to 0.730$ & $0.016 \to 0.291$ & $0.874 \to 0.623$ \\
\kiwi{} & $24.82 \to 27.70$ & $0.888 \to 0.844$ & $0.041 \to 0.052$ & $1.000 \to 1.000$ \\
\wanctrl{} & $16.57 \to 15.99$ & $0.924 \to 0.900$ & $0.035 \to 0.046$ & $0.711 \to 0.647$ \\
\bottomrule
\end{tabular}
\caption{Token-level contribution summary for edited-slot classification.}
\label{tab:token_contrib}
\end{table}

\section{Detailed Analysis of Attention Mechanisms}
\label{sec:appendix_attention}

This appendix provides a comprehensive quantitative analysis of the internal cross-attention behaviors across three distinct architectural backends (Hidden, Query, and Kiwi). Our probing experiments systematically investigate two primary dimensions: \textbf{Token-level Concentration} (the sparsity versus uniformity of attention distributions) and \textbf{Group-level Attention Share} (the dynamic routing of attention among different condition modalities). 

\subsection{Overall Attention Sparsity and Routing Mechanisms}

To evaluate whether the model concentrates its attention on a sparse subset of key condition tokens or distributes it broadly, we compute the Entropy, Gini coefficient, Top-\(K\) Mass (the proportion of total attention weight captured by the top \(K\) tokens), and Head Jaccard@16 (the overlap of top-16 tokens across different attention heads). The results are summarized in Tab.~\ref{tab:overall_concentration}.

\begin{table}[htbp]
\centering
\caption{Overall Attention Concentration and Head Agreement Across Architectural Backends.}
\label{tab:overall_concentration}
\resizebox{\textwidth}{!}{
\begin{tabular}{llccccc}
\toprule
\textbf{Backend} & \textbf{Condition Modality} & \textbf{Entropy} \(\downarrow\) & \textbf{Gini} \(\uparrow\) & \textbf{Top-16 Mass} \(\uparrow\) & \textbf{Top-32 Mass} \(\uparrow\) & \textbf{Head Jaccard@16} \(\uparrow\) \\
\midrule
\textbf{Hidden} & Vision + Text & 3.6658 & 0.9254 & 70.58\% & 84.83\% & 0.3989 \\
\textbf{Query}  & Text + Query  & 3.7143 & 0.9286 & 68.04\% & 83.75\% & 0.3958 \\
\textbf{Kiwi}   & Learnable Queries & 5.6351 & 0.5786 & 13.43\% & 22.88\% & 0.1163 \\
\bottomrule
\end{tabular}
}
\end{table}

The empirical observations reveal a stark contrast in how different architectures route information. The \textbf{Kiwi} backend, which relies on a single group of learnable queries, operates akin to a \textit{distributed memory bank}: it exhibits high entropy, low Top-\(K\) mass, and minimal overlap between attention heads (Jaccard@16 = 0.1163). This indicates that different attention heads explore distinct, non-overlapping subspaces of the queries. Conversely, the multi-group backends (\textbf{Hidden} and \textbf{Query}) exhibit highly sparse distributions. In these architectures, the Top-32 tokens cover over 80\% of the attention mass, and heads show much higher consensus (Jaccard@16 \(\approx\) 0.40). This suggests that combining text with visual or query tokens encourages the network to dynamically route information through a sparse bottleneck of critical condition tokens.

\subsection{Spatial-Temporal Handoff in Multi-Group Architectures}

For architectures containing multiple groups of condition tokens (Hidden and Query), we further analyze the dynamic shift of attention across different denoising steps (temporal) and network layers (spatial). 

\begin{table}[htbp]
\centering
\caption{Dynamic Shift of Group Attention Share in Hidden and Query Backends.}
\label{tab:group_share_dynamics}
\begin{tabular}{llcc}
\toprule
\textbf{Backend} & \textbf{Dimension} & \textbf{Group 1 Share} & \textbf{Group 2 Share} \\
\midrule
\textbf{Hidden} & \textbf{Overall Average} & Vision: 41.35\% & Text: 58.65\% \\
\cmidrule{2-4}
(Vision + Text) & \textbf{Temporal} (Early \(\rightarrow\) Late Steps) & 39.21\% \(\rightarrow\) 43.62\% & 60.79\% \(\rightarrow\) 56.38\% \\
& \textbf{Spatial} (Dual vs. Single Layers) & 36.45\% vs. 46.25\% & 63.55\% vs. 53.75\% \\
\midrule
\textbf{Query} & \textbf{Overall Average} & Text: 57.27\% & Query: 42.74\% \\
\cmidrule{2-4}
(Text + Query) & \textbf{Temporal} (Early \(\rightarrow\) Late Steps) & 61.57\% \(\rightarrow\) 53.09\% & 38.43\% \(\rightarrow\) 46.91\% \\
& \textbf{Spatial} (Dual vs. Single Layers) & 62.19\% vs. 52.34\% & 37.81\% vs. 47.66\% \\
\bottomrule
\end{tabular}
\end{table}

As shown in Tab.~\ref{tab:group_share_dynamics}, we observe a clear ``handoff'' mechanism. Temporally, text tokens predominantly serve as semantic anchors during the early denoising steps, while visual or query tokens incrementally take over the attention share in the later stages. Spatially, the \texttt{Dual} layers heavily favor text semantic alignment, whereas the \texttt{Single} layers shift focus toward visual/query execution. This spatial-temporal handoff implies that the generation process inherently follows a coarse-to-fine paradigm: text establishes the global semantic layout early on and in shallower layers, while visual/query tokens execute fine-grained structural and material editing later in the process.

\subsection{Architectural Determinism Across Editing Attributes}

Finally, we verify the robustness of these attention patterns by evaluating across different editing attributes (Action, Material, Color). 

\begin{table}[htbp]
\centering
\caption{Core Attention Metrics Across Different Editing Attributes.}
\label{tab:attribute_robustness}
\begin{tabular}{llccc}
\toprule
\textbf{Backend} & \textbf{Attribute Type} & \textbf{Core Group Share} & \textbf{Entropy} & \textbf{Top-16 Mass} \\
\midrule
\textbf{Hidden} & Action   & Vision: 41.10\% & 3.7068 & 69.66\% \\
                & Material & Vision: 40.13\% & 3.6309 & 71.41\% \\
                & Color    & Vision: \textbf{42.89\%} & 3.6462 & 70.98\% \\
\midrule
\textbf{Query}  & Action   & Query: 42.21\% & 3.7390 & 67.23\% \\
                & Material & Query: 42.66\% & 3.6862 & 68.92\% \\
                & Color    & Query: \textbf{43.51\%} & 3.7095 & 68.26\% \\
\bottomrule
\end{tabular}
\end{table}

Interestingly, as shown in Tab.~\ref{tab:attribute_robustness}, the attention concentration and group share metrics remain remarkably stable regardless of the specific editing instruction. Although color editing relies slightly more on the Vision/Query groups, the overall variance is marginal. This confirms a principle of \textit{architectural determinism}: the observed routing bottlenecks and attention sparsity are inherent properties of the model's structural design and alignment strategy, rather than artifacts of specific prompt semantics. Consequently, resolving multimodal alignment bottlenecks requires fundamental architectural interventions rather than merely augmenting training data with specific attribute types.

\section{Output Failure Counts}
\label{app:output_counts}

Table~\ref{tab:output_counts} details the absolute counts of various failure modes observed during our evaluation. Note that all percentage rates reported in the main paper are calculated exclusively based on these successfully evaluated outputs.

\begin{table}[h]
\centering
\small
\begin{tabular}{lrrrrrr}
\toprule
\textbf{Interface}  & \textbf{Pass} & \textbf{Under} & \textbf{Wrong slot} & \textbf{Wrong binding} & \textbf{Wrong target} & \textbf{Undet.} \\
\midrule
\uhid{}  & 30 & 15 & 38 & 31 & 3 & 2 \\
\uquery{}  & 31 & 17 & 46 & 40 & 4 & 1 \\
\kiwi{}  & 52 & 20 & 35 & 21 & 6 & 3 \\
\wanctrl{}  & 34 & 15 & 31 & 46 & 8 & 1 \\
\bottomrule
\end{tabular}
\caption{Absolute counts of evaluation outcomes. Each interface was tested on $144$ generated examples. Outputs that were unavailable or rejected by the evaluator are excluded from the rate computations in the main text.}
\label{tab:output_counts}
\end{table}

\subsection{Additional Qualitative Results}
\label{app:qualitative_results}

In this section, we present further qualitative comparisons on complex multi-region video editing tasks, utilizing a $2 \times 2$ grid layout. These highly challenging scenarios evaluate the models' capacity to execute localized edits while preventing ``concept bleeding'' into non-target regions.

\begin{figure}[htbp]
  \centering
  \includegraphics[width=\textwidth]{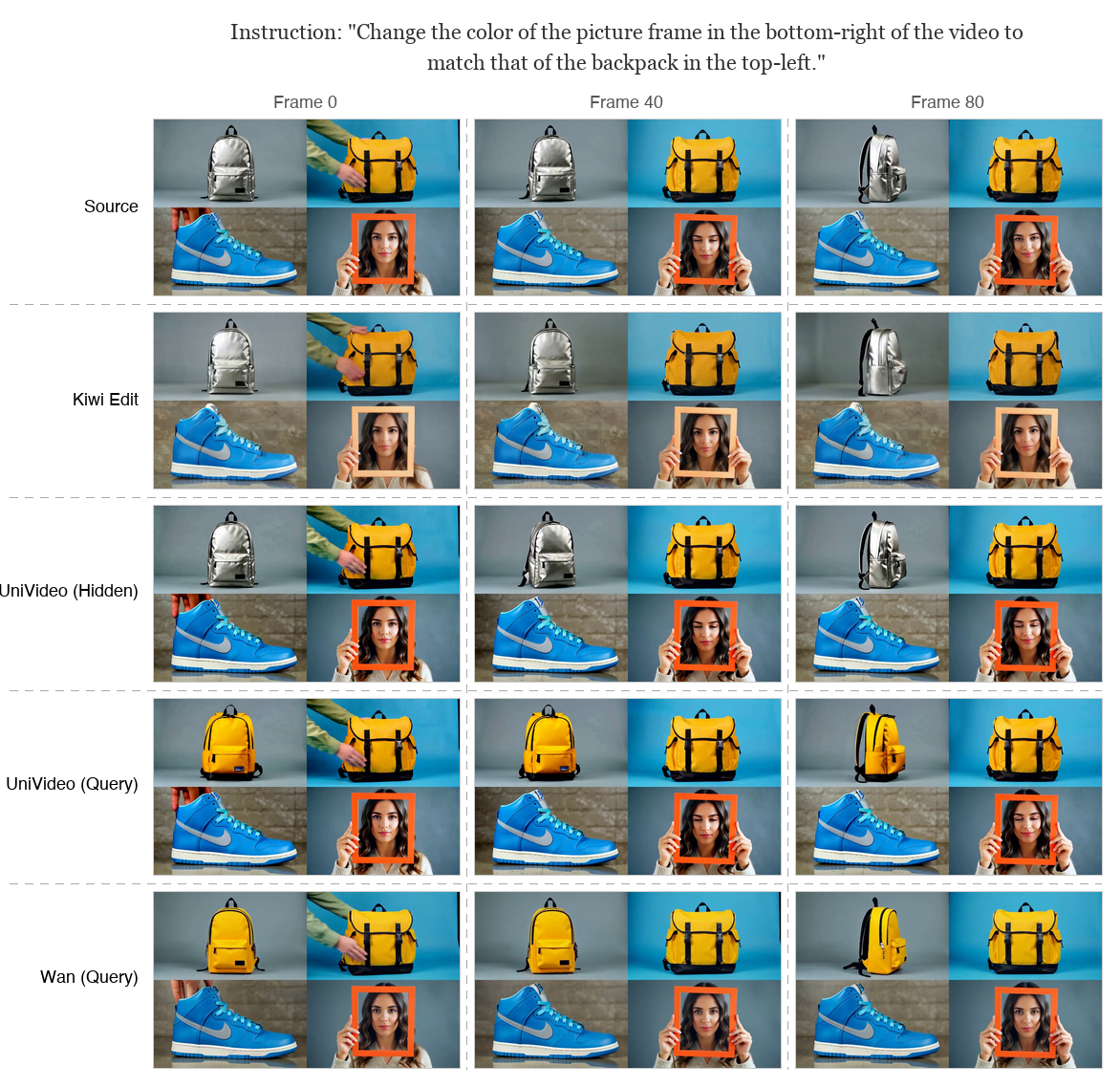}
  \caption{Qualitative comparison of localized editing. The instruction targets the bottom-right picture frame. Both \textit{\uquery{}} and \textit{\wanctrl{}} incorrectly edit the top-left backpack instead, while \textit{\kiwi{}} targets the correct region but fails to achieve the desired silver color.}
  \label{fig:appendix_backpack}
\end{figure}

As illustrated in Fig.~\ref{fig:appendix_backpack}, given the instruction \textit{``Change the color of the picture frame in the bottom-right of the video to match that of the backpack in the top-left,''} the models are required to modify the bottom-right region. However, \textbf{\uquery{}} and \textbf{\wanctrl{}} incorrectly target the wrong region, changing the color of the backpack to yellow while leaving the picture frame unchanged. \textbf{\kiwi{}} correctly targets the bottom-right frame but fails to match the silver color, altering it to a wooden texture instead. Meanwhile, \textbf{\uhid{}} struggles to apply any noticeable edit (under-editing).

\begin{figure}[htbp]
  \centering
  \includegraphics[width=\textwidth]{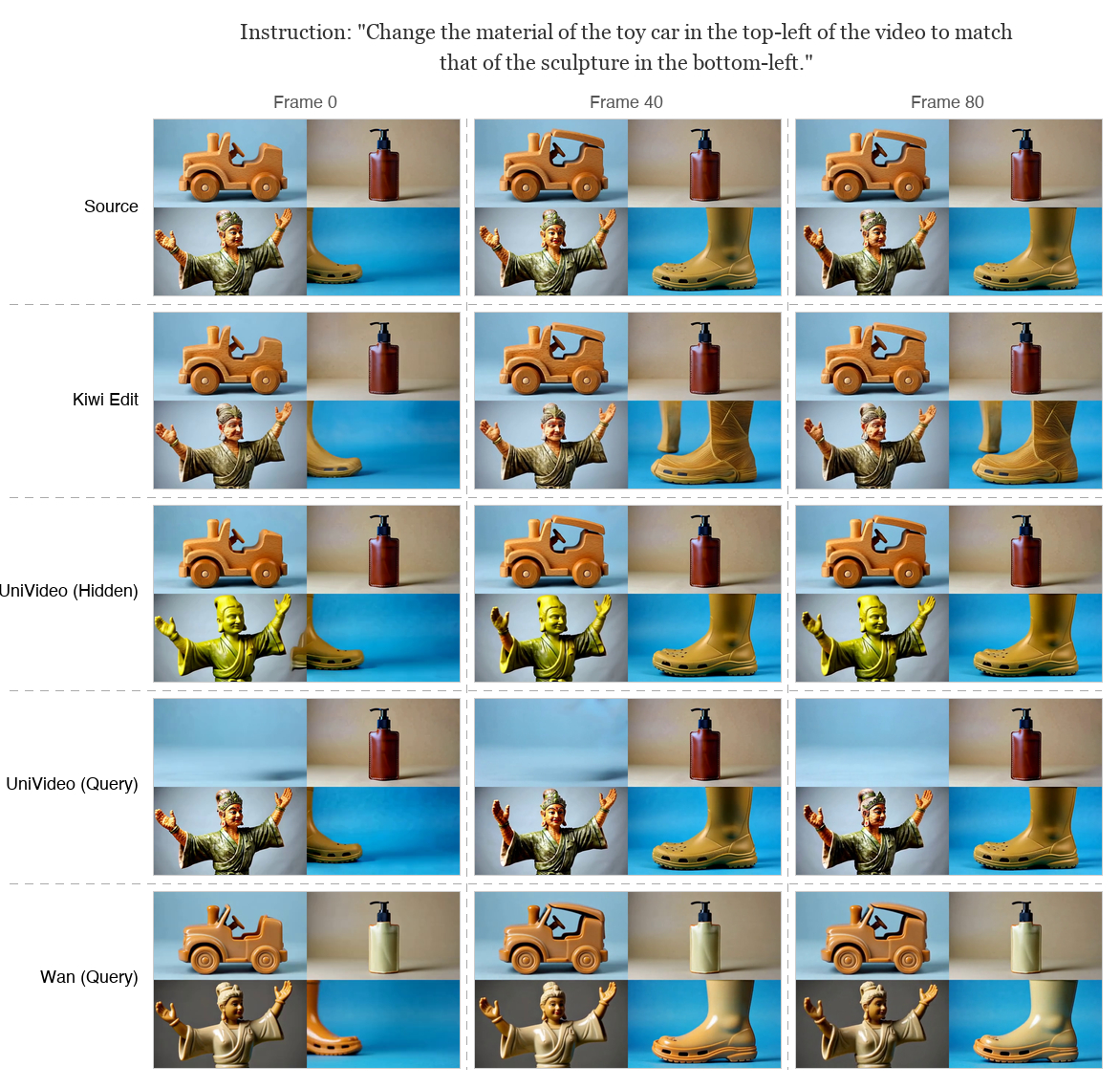}
  \caption{Failure cases in localized material editing. \textit{\uquery{}} suffers from object disappearance (the top-left car vanishes), and \textit{\wanctrl{}} exhibits severe over-editing, altering the materials of non-target objects across all quadrants.}
  \label{fig:appendix_car}
\end{figure}

Fig.~\ref{fig:appendix_car} highlights common failure modes under the instruction \textit{``Change the material of the toy car in the top-left of the video to match that of the sculpture in the bottom-left.''} \textbf{\uquery{}} suffers from severe object disappearance, as the target toy car completely vanishes from the scene. Conversely, \textbf{\wanctrl{}} exhibits extreme over-editing and concept bleeding, drastically altering the materials and colors of all non-target objects (e.g., the sculpture, boot, and bottle) to a uniform ceramic-like texture. Furthermore, \textbf{\uhid{}} incorrectly alters the non-target sculpture instead of the car. These findings underscore the ongoing challenge of balancing precise instruction adherence with the preservation of non-target identities in complex compositional scenes.

% ==========================================
% Figures Inclusion
% ==========================================

\end{document}